\documentclass{article} 
\usepackage{iclr2026_conference,times}

\usepackage{amsmath,amsfonts,bm}

\def\eqref#1{equation~\ref{#1}}

\def\1{\bm{1}}

\DeclareMathAlphabet{\mathsfit}{\encodingdefault}{\sfdefault}{m}{sl}
\SetMathAlphabet{\mathsfit}{bold}{\encodingdefault}{\sfdefault}{bx}{n}

\usepackage[utf8]{inputenc} 
\usepackage[T1]{fontenc}    
\usepackage{hyperref}       
\usepackage{url}            
\usepackage{booktabs}       
\usepackage{amsfonts}       
\usepackage{nicefrac}       
\usepackage{microtype}      
\usepackage{xcolor}         
\usepackage[inline]{enumitem}
\usepackage{graphicx}    
\usepackage{makecell}    
\usepackage{multirow}
\usepackage{overpic}
\usepackage{subfig}
\usepackage{float}
\usepackage{wrapfig}
\usepackage{ragged2e}
\usepackage{xcolor,colortbl}
\usepackage[skins]{tcolorbox}
\usepackage{tikz}
\usepackage{pifont}
\tcbuselibrary{breakable}

\newcolumntype{L}[1]{>{\RaggedRight\arraybackslash}p{#1}}
\newcolumntype{C}[1]{>{\centering\arraybackslash}p{#1}}

\makeatletter
\renewcommand{\paragraph}{%
  \@startsection{paragraph}{4}{\z@}%
                {0.0ex \@plus 0.3ex \@minus 0.1ex}%
                {-1em}%
                {\normalsize\bf}%
}
\makeatother

\def \dataset {{InfoDet}}

\def \eg {{\emph{e.g}.\thinspace}}
\def \numtotal {{101,264}}
\def \numreal {{11,264}}
\def \numsyn  {{90,000}}
\def \numanno {{14,227,680}}
\def \numplat {{10}}

\usepackage{tikz}
\usepackage{xcolor}

\newcommand{\blackcircle}[1]{%
  \tikz[baseline=-1mm]{
    \node[draw=black,
          fill=white,
          circle,
          inner sep=0.2pt,
          text=black,
          font=\footnotesize\rmfamily] (char){#1};}}

\title{\dataset{}: A Dataset for Infographic Element Detection}

\author{Jiangning Zhu$^{1}$, Yuxing Zhou$^{1}$, Zheng Wang$^{1}$, Juntao Yao$^{1}$ \\\textbf{Yima Gu$^{1}$, Yuhui Yuan$^{2}$\hspace{1.5mm}, Shixia Liu}$^{1}$\thanks{Corresponding author.}\\
	$^1$BNRist, Tsinghua University $^2$Canva CORE\\
}

\iclrfinalcopy 

\begin{document}

\maketitle

\begin{figure}[H]
\centering
\includegraphics[width=0.97\linewidth]{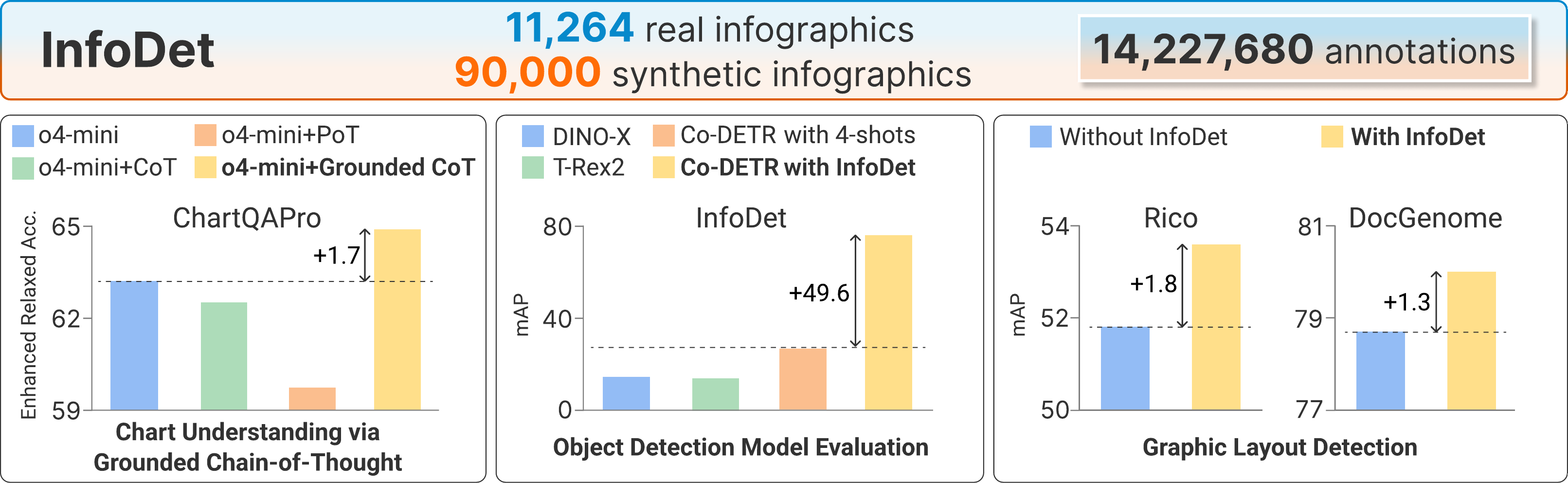}
\caption{\footnotesize Key contributions: 1) An open-source dataset \dataset{}. 2) Improvements on chart understanding, infographic element detection, and graphic layout detection.}
\label{fig:teaser}
\vspace{-4mm}
\end{figure}

\begin{abstract}
Given the central role of charts in scientific, business, and communication contexts, enhancing the chart understanding capabilities of vision-language models (VLMs) has become increasingly critical. 
A key limitation of existing VLMs lies in their inaccurate visual grounding of infographic elements, including charts and human-recognizable objects (HROs) such as icons and images.
However, chart understanding often requires identifying relevant elements and reasoning over them.
To address this limitation, we introduce \dataset{}, a dataset designed to support the development of accurate object detection models for charts and HROs in infographics.
It contains $\numreal{}$ real and $\numsyn{}$ synthetic infographics, with over 14 million bounding box annotations. 
These annotations are created by combining the model-in-the-loop and programmatic methods.
We demonstrate the usefulness of \dataset{} through three applications: 1) constructing a Thinking-with-Boxes scheme to boost the chart understanding performance of VLMs, 2) comparing existing object detection models, and 3) applying the developed detection model to document layout and UI element detection.

\raisebox{-0.3\height}{\hspace{0.05cm}\includegraphics[width=0.45cm]{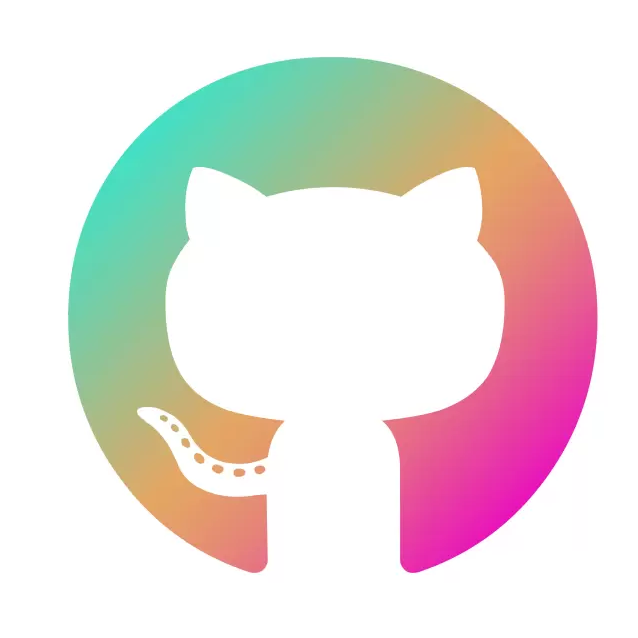}} \small \textbf{\mbox{Code:}} \href{https://github.com/InfoDet2025/InfoDet}{https://github.com/InfoDet2025/InfoDet} \\
\vspace{1em}
\raisebox{-0.3\height}{\includegraphics[width=0.4cm]{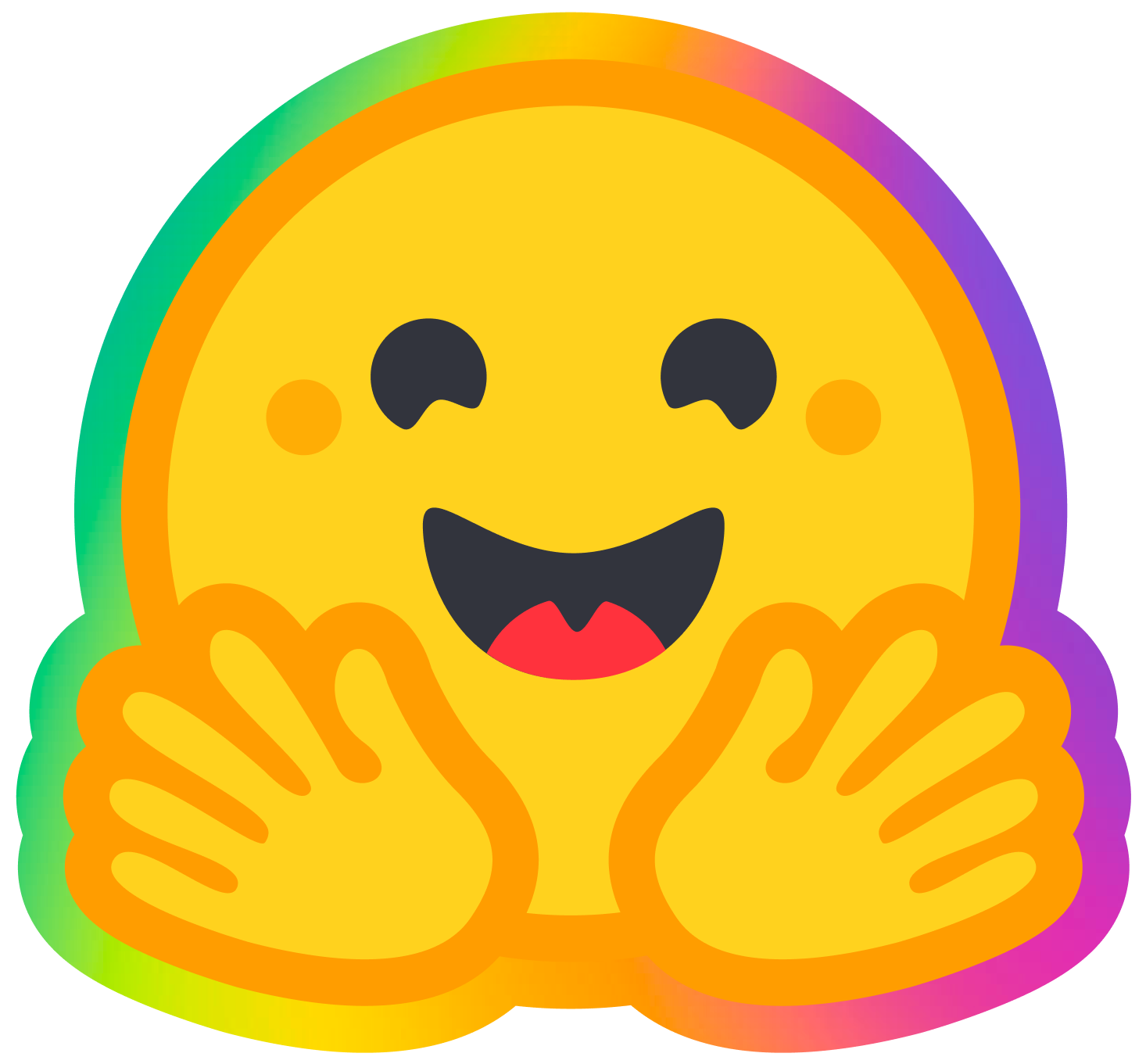}} \small \textbf{\mbox{Data \& Dataset Card:}} \href{https://huggingface.co/datasets/InfoDet/InfoDet}{https://huggingface.co/datasets/InfoDet/InfoDet}

\end{abstract}

\section{Introduction}

Charts are a fundamental medium for conveying data-driven insights across scientific, business, and communication domains.
Consequently, improving vision-language models (VLMs) for chart understanding has become increasingly critical, driving significant advances in understanding plain charts~\citep{huang2024pixels}--minimal combinations of texts and charts.
In practice, however, charts are often combined with icons and images of real-world objects, known as human-recognizable objects (HROs)~\citep{borkin2015beyond}, to create infographics.
By thoughtfully arranging texts, charts, and HROs, infographics transform abstract data into accessible insights through engaging visual designs.
While effective for human interpretation, these designs introduce difficulties for VLMs in accurately understanding chart content~\citep{masry2025chartqapro}.
Previous studies~\citep{masry2025chartqapro,li2025trigbench} have identified a key limitation of existing VLMs: inaccurate visual grounding of infographic elements, which hinders the ability to associate the elements with the underlying data.
This highlights the need for robust object detection models to support visual grounding and enhance chart understanding.
Although considerable progress has been made in text detection~\citep{long2021scene,du2020pp}, detecting charts and HROs---key elements linking abstract data to human perception---remains relatively underexplored.

Compared to natural scenes, element detection in infographics presents challenges for two reasons.
First, infographic elements exhibit high intra-class variance.
Charts vary widely in type, layout, and visual design, and HROs appear in diverse styles, spanning from realistic depictions to abstract representations of real-world objects. 
Second, the visual interplay between charts and HROs often results in ambiguous boundaries, making it difficult to distinguish one element from another in context.
To effectively handle the highly varied infographic elements with ambiguous boundaries, the detection model needs to learn from a diverse set of infographics with accurate annotations.
Existing datasets, however, primarily focus on plain charts without HROs~\citep{battle2018beagle,deng2022visimages}, failing to capture the complexity of infographics.
\citet{borkin2015beyond} have taken the first step in building a dataset with rich annotations, but their dataset is limited in scale, comprising only $393$ samples due to the labor-intensive manual annotation process.
To advance element detection in infographics, a large-scale dataset of diverse infographics with comprehensive annotations is required.

To fill this gap, we introduce \dataset{}, a dataset for infographic element detection.
It comprises a diverse collection of infographics from two sources: 1) real infographics collected from \numplat{} online platforms, such as \citet{visualcapitalist} and \citet{statista}, and 2) synthetic infographics programmatically created from $1,072$ design templates.
To effectively annotate the infographics, we combine programmatic and model-in-the-loop~\citep{kirillov2023segment} methods.
For the synthetic infographics, we programmatically derive the bounding boxes.
For the real infographics, we co-develop an object detection model and the annotation process, allowing the model and the annotations to iteratively enhance each other.
Specifically, we use the annotated synthetic infographics to fine-tune an InternImage-based object detection model~\citep{wang2023internimage},
which is then employed to generate annotations for all real infographics.
The generated annotations are reviewed and corrected by the experts through multiple rounds of refinement.
In each round, expert feedback is utilized to enhance the annotation quality and refine the model, thereby progressively improving its accuracy.
In total, \dataset{} contains \textbf{\numreal{}} real and \textbf{\numsyn{}} synthetic infographics, along with \textbf{\numanno{}} bounding box annotations of texts, charts, HROs, and finer-grained sub-elements such as bars, axes, and
legends.
Table~\ref{tab:dataset_stats} in the Appendix provides a statistical comparison of \dataset{} with existing chart datasets.

We demonstrate the usefulness of \dataset{} through three applications (Fig.~\ref{fig:teaser}).
First, we propose a Thinking-with-Boxes scheme that performs grounded chain-of-thought reasoning over elements.
This grounded reasoning considerably improves the performance of OpenAI o4-mini on the challenging ChartQAPro benchmark~\citep{masry2025chartqapro}.
Second, we compare the performance of the state-of-the-art object detection models.
The results show that the best-performing foundation models for object detection (\eg, DINO-X~\citep{ren2024dinoxunifiedvisionmodel}) still struggle to accurately detect infographic elements, whereas fine-tuning traditional object detection models (\eg, Faster R-CNN~\citep{ren2015faster}) with \dataset{} achieves improved performance.
These findings highlight the importance of sufficient high-quality training data for chart and HRO detection.
Third, we apply our InternImage-based object detection model to out-of-domain graphic layout detection tasks, including document layout and UI element detection, demonstrating its generalization capability across broader domains.

The main contributions of this work are threefold:
\begin{itemize}
    \setlength\itemsep{0pt}
    \setlength\itemindent{-6mm}
    \item A large-scale dataset for infographic element detection with $\numtotal{}$ annotated infographics.
    \item An InternImage-based model for detecting charts and HROs in infographics. 
    \item Three applications that show \dataset{}'s usefulness in chart understanding and object detection.
\end{itemize}

\section{Related Work}
Based on the presence of HROs, chart datasets with element annotations can be classified into two categories: datasets of plain charts and datasets of infographic charts.

\textbf{Plain charts} present data in a minimal manner using texts and charts.
Some datasets consist of programmatically created charts.
FigureQA~\citep{ebrahimi2018figureqa} comprises $100,000$ charts created from randomly generated data using \citet{bokeh}.
However, relying on randomly generated data limits real-world representativeness.
To address this, \citet{methani2020plotqa} use crawled data to create $224,377$ charts by randomly combining design parameters such as marker and line styles.
Other datasets are constructed by collecting charts from existing literature or online platforms.
VG-DCU~\citep{Dou2024Hierarchically} consists of $15,197$ SVG-based charts, from which bounding box annotations are extracted by analyzing the SVG elements.
VisImages~\citep{deng2022visimages} is constructed by gathering $12,267$ images from IEEE VIS publications and manually annotating $35,096$ charts within them.
While these datasets facilitate object detection model training for plain charts, such models often struggle with the widely used infographics, where diverse HROs and their interplay with the charts introduce significant variability.

To better support the analysis of \textbf{infographic} designs, \citet{borkin2015beyond} pioneered the creation of an infographic dataset with rich annotations.
They utilize an existing database of real infographics and manually annotate the polygons of their elements.
However, this dataset is limited in scale, comprising only $393$ samples due to the labor-intensive manual annotation process.
As a result, this dataset is unsuitable for training object detection models that require strong generalization.
In contrast, \dataset{} combines a model-in-the-loop annotation method for real infographics and a programmatic annotation method for synthetic infographics, resulting in $\numtotal{}$ annotated infographics that effectively support object detection model development.

\begin{figure}[!b]
\centering
\includegraphics[width=0.97\linewidth]{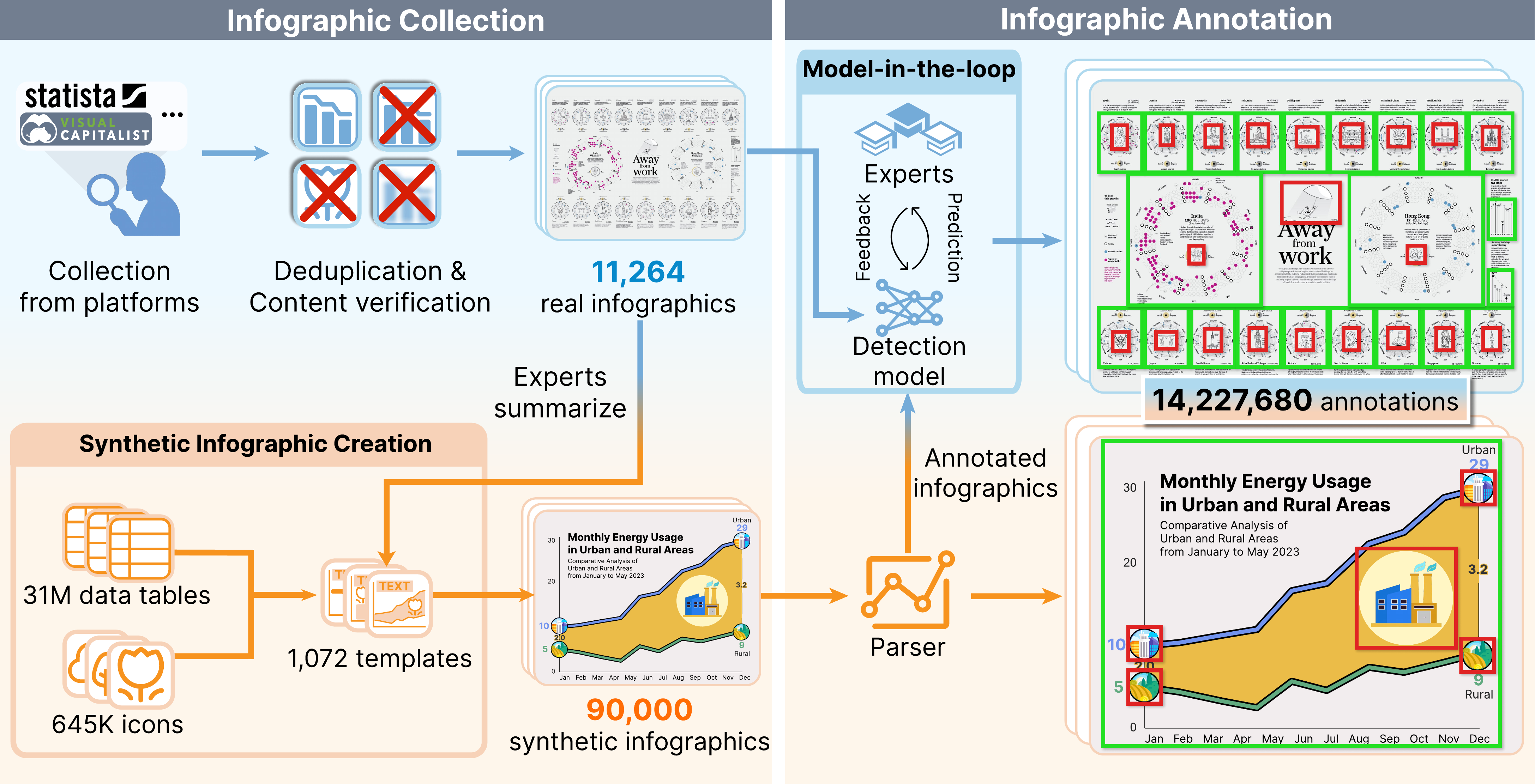}
\caption{The construction pipeline for the \dataset{} dataset.}
\label{fig:pipeline}
\end{figure}

\section{\dataset{} Construction Method}
\label{sec:method}

Fig.~\ref{fig:pipeline} provides an overview of the dataset construction pipeline, which includes two main steps: infographic collection and infographic annotation.

\subsection{Infographic Collection}

Previous studies~\citep{zhu2019towards,zhu2025reorderbench} have highlighted the complementary benefits of real and synthetic data: the former captures authentic design practices, while the latter offers controlled variation and scalability for robust training and evaluation.
Informed by this finding, we collect infographics from two sources to balance authenticity, diversity, and scalability:
\begin{enumerate*}[label={\arabic*)}]
    \item real infographics from chart-rich online platforms, and
    \item synthetic infographics created programmatically
\end{enumerate*}.

\paragraph{Real infographic collection}
We collect infographics from \numplat{} chart-rich online platforms that permit research use, such as \citet{visualcapitalist} and \citet{statista}.
The complete list is provided in Appendix~\ref{appx:platforms}.
To enhance the quality of the infographics, we implement a two-step filtering process: deduplication and content verification.
In deduplication, we remove infographics that exhibit high CLIP similarity~\citep{radford2021learning} ($\geq 0.9$) and low perceptual hashing distance~\citep{idealods2019imagededup} ($\leq 2$) relative to other infographics.
In content verification, we prompt GPT-4o mini to identify and remove images that are blurry, lack charts or HROs, are photographs instead of graphic designs, or contain personal or sensitive content.
After filtering, the collection is refined to $\numreal{}$ high-quality infographics.

\paragraph{Synthetic infographic creation}
We employ a template-based method~\citep{li2025chartgalaxy} to create synthetic infographics.
This method utilizes $1,072$ design templates derived from representative real infographics.
Each template specifies:
\begin{enumerate*}[label={\arabic*)}]
    \item the presence and relative positions of charts, texts, and HROs, and
    \item the type and visual style of the charts.
\end{enumerate*}
An infographic is created by filling the template with:
\begin{enumerate*}[label={\arabic*)}]
    \item data tables for chart creation, 
    \item descriptive texts, and
    \item selected HROs 
\end{enumerate*}.
To ensure diversity, we sample data tables from VizNet~\citep{hu2019viznet}, a large-scale dataset containing over 31 million tables and associated metadata. 
Charts are created from the sampled data tables as specified by the template.
Descriptive texts for the charts are generated using GPT-4o mini.
HROs with the highest CLIP similarity to the descriptive texts are selected from the IconQA dataset~\citep{lu2021iconqa}, which contains over 645K icons.
Using this template-based method, we generate $\numsyn{}$ synthetic infographics.
Example templates and infographics generated from them are provided in Appendix~\ref{app:example}.

\subsection{Infographic Annotation}

Given the differences in collecting real and synthetic infographics, we adopt two annotation methods: a programmatic method for synthetic infographics and a model-in-the-loop method for real ones.
\looseness=-1

\paragraph{Programmatic synthetic infographic annotation}
Synthetic infographic annotations are programmatically generated with a parser integrated into the infographic generation process. 
This parser extracts bounding boxes for texts, charts, and HROs from the corresponding SVG file, which encodes the visual and structural details of the infographic.
Additionally, the parser leverages information from the design template to classify charts and HROs into subcategories: charts are categorized into $75$ distinct types, while HROs are labeled as either data-related or theme-related objects.
The complete list of chart types is provided in Appendix~\ref{app:stats}.

\paragraph{Model-in-the-loop real infographic annotation}
To reduce human labor in the annotation, we aim to leverage object detection models for assistance.
However, there is an absence of specialized detection models for charts and HROs.
To address this, we employ a model-in-the-loop annotation method~\citep{kirillov2023segment}.
This method co-develops an object detection model and the annotation process, allowing the model and the annotations to iteratively enhance each other.
Specifically, using the annotated synthetic infographics, we build an object detection model by fine-tuning InternImage-L~\citep{wang2023internimage} along with the DINO~\citep{zhang2023dino} detector.
This fine-tuned model is then employed to generate annotations for all real infographics.
However, since the synthetic infographics do not fully represent the diversity of all infographics, the fine-tuned object detection model is prone to errors.
To mitigate this, we conduct multiple rounds of annotation refinement and model enhancement with the experts.
In each round, the experts review and correct the auto-generated annotations, and the feedback is used to further fine-tune the model, progressively improving its accuracy.
At the end of the refinement process, we randomly sample $1,250$ infographics to evaluate the quality of the generated annotations.
Results show that the generated annotations achieve a precision of $93.9\%$ and a recall of 
$96.7\%$, comparable to those of widely used object detection datasets, such as COCO~\citep{lin2014microsoft} ($83.0\%$ recall and $71.9\%$ precision) and Objects365~\citep{shao2019objects365} ($92.0\%$ recall and $91.7\%$ precision), as reported by~\citet{shao2019objects365}.

\subsection{Statistics and Dataset Analysis}

\dataset{} contains \textbf{\numtotal{}} infographics, including \textbf{\numreal{}} real and \textbf{\numsyn{}} synthetic infographics.
To complement the dataset with text annotations, we use the widely adopted OCR model PP-OCRv4~\citep{du2020pp} to annotate all real infographics and extract text annotations from the generation process for synthetic infographics.
For completeness, we also extract mark-level annotations of $26$ categories of chart sub-elements, such as bars, axes, and legends.
The list of sub-element categories is provided in Appendix~\ref{app:mark_cats}.
Across these infographics, we annotate a total of \textbf{7,284,892} text elements (each corresponding to a line of text), \textbf{310,299} charts, \textbf{1,080,598} HROs, and \textbf{5,551,891} sub-elements.
The detailed statistics are provided in Appendix~\ref{app:stats}.
Beyond the basic statistics, we have verified that our dataset exhibits high diversity, no harmful bias, and high fidelity of synthetic infographics (See Appendix~\ref{app:dataset_analysis}).
To ensure consistent evaluation, we split \dataset{} into a training set of $96,264$ infographics and a test set of $5,000$ infographics, while maintaining the same proportion of real and synthetic infographics in both sets.

\section{Experiments}
\label{sec:experiment}

In this section, we first construct a Thinking-with-Boxes scheme to enhance the performance of the latest VLMs.
We then evaluate the performance of existing object detection models.
Finally, we apply the InternImage-based object detection model to graphic layout detection tasks.

\begin{figure}[!b]
\centering
\begin{overpic}[width=0.97\linewidth]{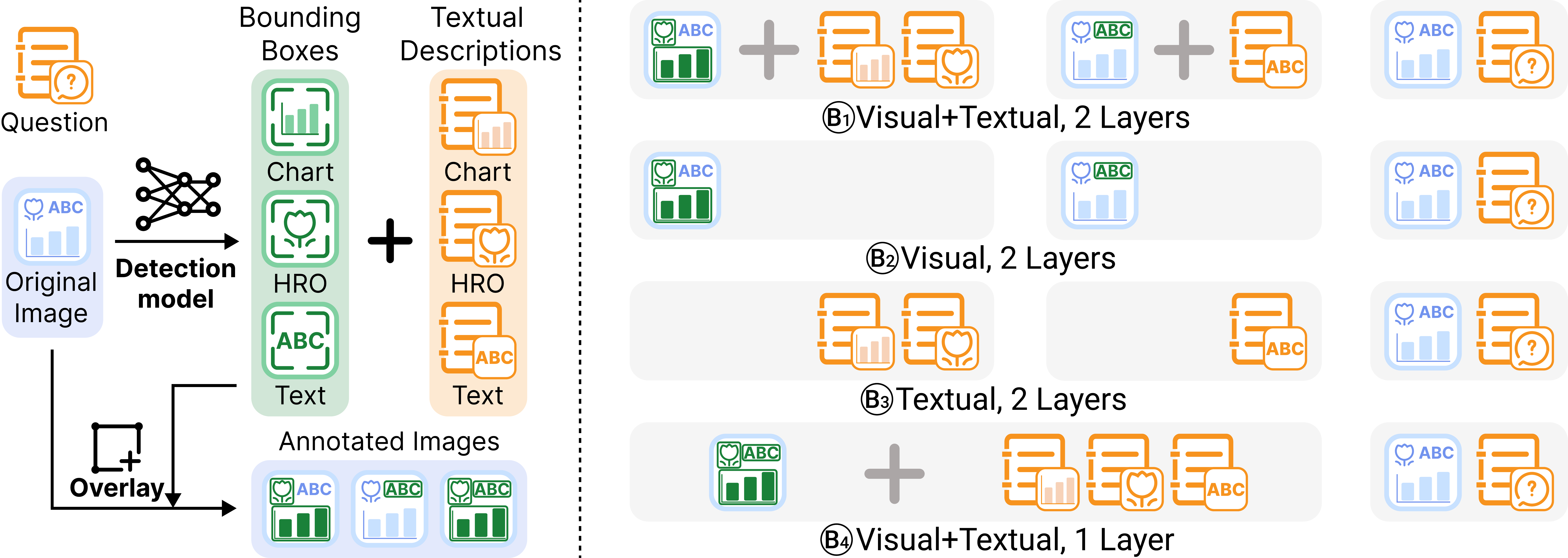}
\end{overpic}
\put(-320, -12){(a)}
\put(-127, -12){(b)}
\caption{The Thinking-with-Boxes scheme: (a) the charts, HROs, and texts are detected and overlaid onto the original image to create annotated images with grounded elements; (b) the input of the grounded chain-of-thought method (B$_1$) and its ablated variants (B$_2$, B$_3$, B$_4$).}
\label{fig:gcot_ablate}
\end{figure}

\subsection{Thinking-with-Boxes via Grounded Chain-of-Thought}

Latest VLMs, such as OpenAI's \citet{o3}/\citet{o4-mini}, demonstrate chain-of-thought reasoning capability with images through seamless image manipulations, including automatic zooming and cropping.
Considering that chart understanding requires more complex, fine-grained visual reasoning over the elements within infographic images~\citep{lin2025infochartqa}, we construct a Thinking-with-Boxes scheme to enhance VLMs by explicitly providing grounded annotations of texts, charts, and HROs along with additional layered infographic images. 
The bounding boxes are predicted using our infographic-oriented object detection model and an OCR model.
With this scheme, we prompt the VLMs to output reasoning trajectories over the grounded regions, referred to as \emph{grounded chain-of-thought} (grounded CoT), which guide the model to think step-by-step before achieving the final answer.
Next, we detail the implementation of grounded CoT and demonstrate the effectiveness of the Thinking-with-Boxes scheme through improved performance on ChartQAPro.

\subsubsection{Grounded Chain-of-Thought Prompting}

As shown in Fig.~\ref{fig:gcot_ablate}\blackcircle{B$_1$}, we provide the VLM with detected elements in two modalities—visual prompts, by overlaying boxes on the infographic image, and textual descriptions of each element—to study the reasoning preferences of the evaluated VLMs.

For the \textbf{visual prompts}, we overlay bounding boxes on top of the infographics, each labeled with an alphabetical ID.
To improve clarity, the bounding boxes are rendered in contrastive colors against the background, and the ID labels are placed to minimize overlap.
However, even with these measures, overlap between bounding boxes remains inevitable in regions with dense texts and HROs.
To mitigate this, we propose to separate the visual prompts into two layers: one containing charts and HROs, and the other containing texts. 
We also provide \textbf{textual description} of each element to ease the challenge of simultaneously locating and interpreting their content.
Please refer to Appendix~\ref{app:GCoT_setup} for detailed prompts and a comparison of visual prompts rendered in one versus two layers.

\begin{table}[!t]
  \centering
  \caption{Performance of o1, o3, and o4-mini with different prompting methods. The best one is \textbf{bold}.}
  \resizebox{0.95\textwidth}{!}{
    \renewcommand{\arraystretch}{1.0} 
    \begin{tabular}{l|cccc|cccc|cccc}
    \toprule
    \multirow{2}[3]{*}{\textbf{Chart Group}} & \multicolumn{4}{c|}{\textbf{\makecell{o1}}} & \multicolumn{4}{c|}{\textbf{\makecell{o3}}} & \multicolumn{4}{c}{\textbf{\makecell{o4-mini}}} \\
    & Direct & CoT & PoT & \makecell{Grounded\\CoT (ours)} & Direct & CoT & PoT & \makecell{Grounded\\CoT (ours)} & Direct & CoT & PoT & \makecell{Grounded\\CoT (ours)}\\
    \midrule
    Plain, Single & 57.8 & 57.8 & 56.1 & \textbf{60.1} & 56.8 & \textbf{57.7} & 57.5 & 57.2 & 58.1 & 57.9 & 55.3 & \textbf{60.6} \\
    Plain, Multiple & 63.7 & 65.1 & 62.2 & \textbf{65.4} & 62.8 & 61.0 & 58.8 & \textbf{63.4} & 66.7 & 66.1 & 62.3 & \textbf{66.9} \\
    Infographic, Single & 66.4 & 64.3 & 60.9 & \textbf{67.8} & 64.9 & 59.5 & 64.2 & \textbf{67.7} & 67.4 & 64.4 & 67.5 & \textbf{68.4} \\
    Infographic, Multiple & 66.0 & 67.6 & 66.8 & \textbf{71.9} & 66.0 & 64.9 & 64.2 & \textbf{68.8} & 70.6 & 69.2 & 64.7 & \textbf{72.5} \\
    \midrule
    Overall & 61.4 & 61.9 & 60.0 & \textbf{64.1} & 60.6 & 60.0 & 59.5 & \textbf{61.6} & 63.2 & 62.5 & 59.7 & \textbf{64.9} \\
    \bottomrule
    \end{tabular}
  }
  \label{tab:o1o3o4mini_eval}%
\end{table}

\begin{figure}[!b]
\centering
\includegraphics[width=0.95\linewidth]{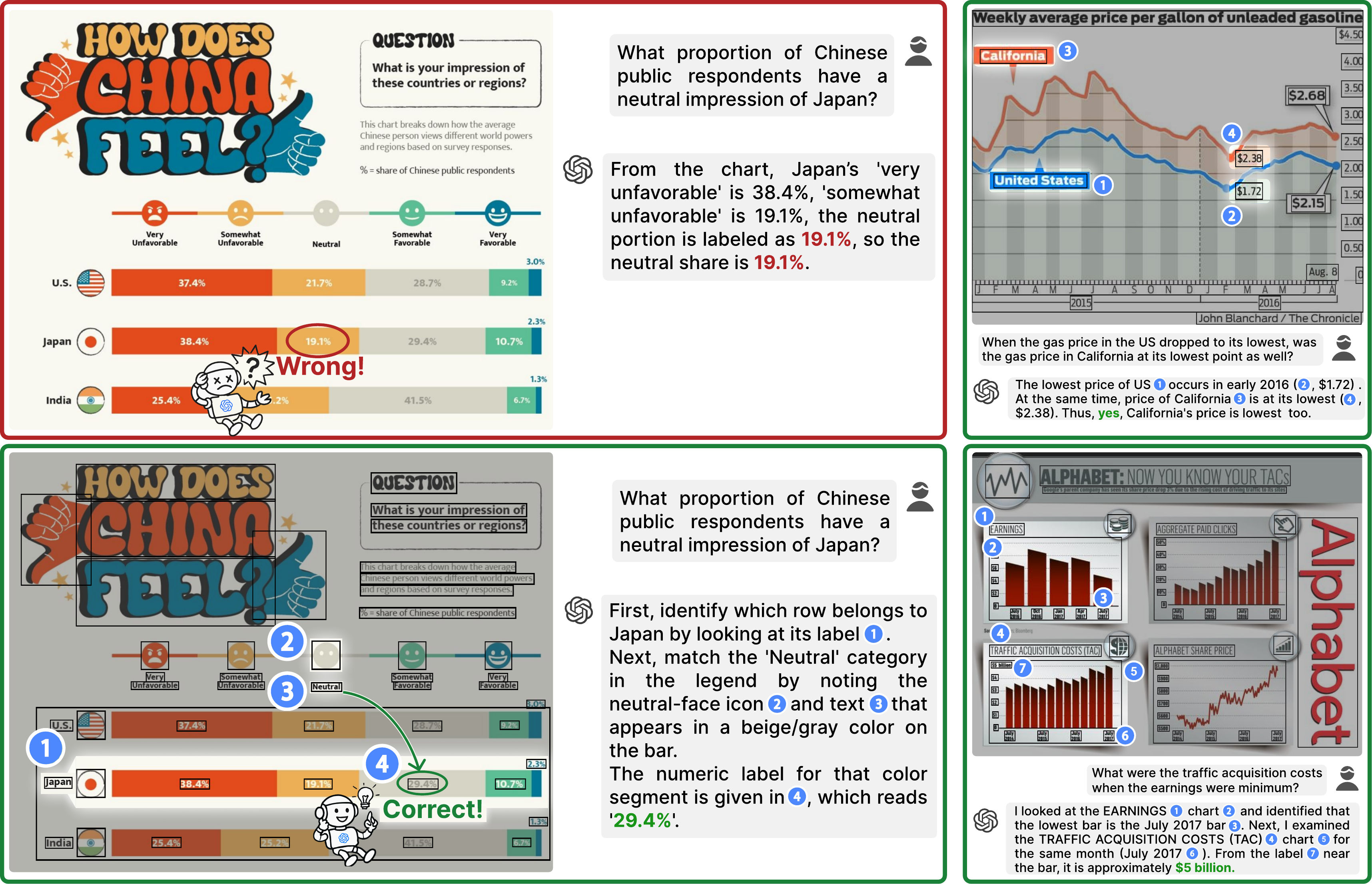}
\caption{Grounded CoT guides the model to think step-by-step and achieve the correct answer.}
\label{fig:gcot_qual}
\end{figure}

\subsubsection{Experimental Setup}

We evaluate the chart understanding capability of VLMs using the ChartQAPro benchmark~\citep{masry2025chartqapro}, which contains $1,948$ challenging question-answer pairs across $1,341$ images.
To better analyze the performance of our method, we manually categorize them into four groups based on two criteria: whether the charts are \textbf{plain} or \textbf{infographic}, and whether there are \textbf{single} or \textbf{multiple} charts.
We assess three state-of-the-art VLMs: OpenAI's \citet{o1}, \citet{o3}, and \citet{o4-mini}.
For each VLM, we compare our method against three widely used baseline prompting methods:
\begin{enumerate*}[label={\arabic*)}]
    \item \textbf{Direct} prompting with the chart image and the question,
    \item \textbf{Chain-of-Thought}~\citep{wei2022chain} (CoT), which prompts the model to reason step-by-step for the provided image and question, and
    \item \textbf{Program-of-Thought}~\citep{chen2023program} (PoT), which prompts the model to generate a Python code that prints the final answer.
\end{enumerate*}
The performance is measured using the enhanced relaxed accuracy~\citep{masry2025chartqapro}. 
Please refer to Appendix~\ref{app:GCoT_setup} for the detailed prompts and the enhanced relaxed accuracy implementation.

\subsubsection{Results and Analysis}

\paragraph{Effectiveness of grounded CoT prompting}
As shown in Table~\ref{tab:o1o3o4mini_eval}, prompting the latest VLMs to think step-by-step or write Python code does not significantly improve their performance.
We attribute this to the reasoning-centric design of the VLMs, which inherently reduces the dependence on explicit prompts for step-by-step reasoning.
In contrast, our method enhances chart understanding performance by providing grounded infographic elements. 
In particular, our method performs comparably on plain, single charts and shows better performance on infographic charts and images with multiple charts, leading to improved overall performance.
As shown in Fig.~\ref{fig:gcot_qual}, the grounded annotations of elements effectively guide the VLM to reason step-by-step and arrive at the correct answer.
Despite its strong visual reasoning capability, o3 encounters instruction-following issues, resulting in slightly lower performance compared to o1 and o4-mini. 
The detailed evaluation results and analysis of this issue are provided in Appendix~\ref{app:GCoT_ana}.

\paragraph{Ablation Study}
We conduct ablation studies on o1 to evaluate the effects of different prompt modality designs, separating grounded annotations into two layers, and adding in-context examples.

\textit{Prompt modality}.
Table~\ref{tab:o1_eval}(a) shows that using only visual prompts (Fig.~\ref{fig:gcot_ablate}\blackcircle{B$_2$}) or textual descriptions (Fig.~\ref{fig:gcot_ablate}\blackcircle{B$_3$}) results in a performance drop compared to combining both.
This highlights their complementary roles in grounding infographic elements and supporting VLMs in chart understanding.

\textit{Prompt separation}. 
Table~\ref{tab:o1_eval}(b) shows that separating the prompts into two layers leads to better performance than providing them in one layer (Fig.~\ref{fig:gcot_ablate}\blackcircle{B$_4$}).
This suggests that reducing overlap through separation facilitates the visual grounding of infographic elements and improves chart understanding.

\textit{Incorporation of in-context examples}. 
Table~\ref{tab:o1_eval}(c) shows that incorporating in-context examples results in a performance drop.
This indicates that the latest VLMs can perform reasoning tasks effectively without additional examples, which would instead introduce confusion and hinder performance.

\begin{table}[!t]
  \centering
  \caption{Ablation of the grounded CoT method. The best one is \textbf{bold}.}
  \subfloat[Prompt Modality]{
    \resizebox{0.3\textwidth}{!}{
      \begin{tabular}{ccc}
        \toprule
        \textbf{Visual} & \textbf{Textual}  & \textbf{Visual+Textual} \\
        \midrule
        62.8 & 61.6 & \textbf{64.1} \\
        \bottomrule
      \end{tabular}
    }
  }
  \subfloat[Prompt Separation]{
    \resizebox{0.2\textwidth}{!}{
      \begin{tabular}{>{\centering\arraybackslash}p{1.4cm}>{\centering\arraybackslash}p{1.4cm}}
        \toprule
        \multicolumn{1}{c}{\textbf{1-Layer}} & \textbf{2-Layer} \\
        \midrule
        62.3 & \textbf{64.1} \\
        \bottomrule
      \end{tabular}
    }
  }
  \subfloat[In-Context Example]{
    \resizebox{0.309\textwidth}{!}{
      \begin{tabular}{cc}
        \toprule
        \textbf{With Example} & \textbf{Without Example} \\
        \midrule
        61.5 & \textbf{64.1} \\
        \bottomrule
      \end{tabular}
    }
  }
  \label{tab:o1_eval}%
  \vspace{-4mm}
\end{table}

\subsection{Evaluating Object Detection Models}

We evaluate the infographic element detection performance of 11 object detection models on \dataset{}.

\subsubsection{Experimental Setup}

\paragraph{Models}
Existing object detection models can be classified into two categories: foundation models that support zero-/few-shot detection and traditional deep learning models that require fine-tuning before detecting novel classes.
We select the representative models in each category, including seven foundation models (RegionCLIP~\citep{zhong2022regionclip}, Detic~\citep{zhou2022detecting}, Grounding DINO~\citep{liu2024grounding}, GLIP~\citep{li2022grounded}, MQ-GLIP~\citep{xu2023multi}, T-Rex2~\citep{jiang2024t}, and DINO-X~\citep{ren2024dinoxunifiedvisionmodel}) and four traditional models (Faster R-CNN~\citep{ren2015faster}, YOLOv3~\citep{redmon2018yolov3}, RTMDet~\citep{lyu2022rtmdet}, and Co-DETR~\citep{zong2023detrs}).

\paragraph{Evaluation protocol}
The above models are not tailored to detecting charts and HROs. 
To address this, we evaluate three adaptation methods:
\begin{enumerate*}[label={\arabic*)}]
    \item \textbf{Zero-shot prompting}, which uses text prompts to define target classes,
    \item \textbf{Few-shot prompting}, which uses $k$ randomly selected infographics to describe target classes, optionally augmented with text prompts, and
    \item \textbf{Standard fine-tuning}, which updates model weights using annotated infographics, either with $k$ random example infographics or the entire \dataset{} training set.
\end{enumerate*}
We compare the models at two granularity levels: the element level targeting charts and HROs, and the mark level targeting sub-elements such as bars, axes, and legends.
For both levels, the performance is measured using the average precision (AP) and recall (AR) on the \dataset{} test set.
Please refer to Appendix~\ref{app:benchmark_setup} for more details on text prompts, fine-tuning hyperparameters, and computational costs.

\subsubsection{Results and Analysis}

\definecolor{foundation_models}{RGB}{185, 235, 255}
\definecolor{traditional_models}{RGB}{245,214,244}

\begin{table}[!t]
  \centering
  \caption{Evaluation results of the \colorbox{foundation_models!70}{foundation} and the \colorbox{traditional_models!70}{traditional} models at the element level. The best one is \textbf{bold}.}
  \subfloat[Zero-shot prompting]{
  \resizebox{0.48\textwidth}{!}{
    \begin{tabular}{p{2.5cm}|>{\centering\arraybackslash}p{1.5cm}>{\centering\arraybackslash}p{1.5cm}|>{\centering\arraybackslash}p{1.5cm}>{\centering\arraybackslash}p{1.5cm}}
    \toprule
    \multirow{2}[1]{*}{\textbf{Model}} & \multicolumn{2}{c|}{\textbf{Average Precision (AP)}} & \multicolumn{2}{c}{\textbf{Average Recall (AR)}} \\
    & Chart & HRO & Chart & HRO \\
    \midrule
    \rowcolor{foundation_models!25} RegionCLIP & 0.8 & 3.6 & 13.9 & 24.9 \\
    \rowcolor{foundation_models!25} Detic & 1.8 & 4.4 & 23.7 & 11.3 \\
    \rowcolor{foundation_models!25} Grounding DINO & 12.6 & 12.2 & \textbf{63.2} & \textbf{46.0} \\
    \rowcolor{foundation_models!25} GLIP & 13.5 & 11.2 & 44.9 & 33.2 \\
    \rowcolor{foundation_models!25} MQ-GLIP & 13.5 & 11.2 & 44.9 & 33.2 \\
    \rowcolor{foundation_models!25} DINO-X & \textbf{14.0} & \textbf{15.0} & 29.4 & 29.1 \\
    \bottomrule
    \end{tabular}%
  }
  }
  \subfloat[Few-shot prompting, 4-shots]{
  \resizebox{0.48\textwidth}{!}{
    \begin{tabular}{p{2.5cm}|>{\centering\arraybackslash}p{1.5cm}>{\centering\arraybackslash}p{1.5cm}|>{\centering\arraybackslash}p{1.5cm}>{\centering\arraybackslash}p{1.5cm}}
    \toprule
    \multirow{2}[1]{*}{\textbf{Model}} & \multicolumn{2}{c|}{\textbf{Average Precsion (AP)}} & \multicolumn{2}{c}{\textbf{Average Recall (AR)}} \\
    & Chart & HRO & Chart & HRO \\
    \midrule
    \rowcolor{foundation_models!25} MQ-GLIP & \textbf{16.2} & 15.5 & \textbf{43.5} & \textbf{40.7} \\
    \rowcolor{foundation_models!25} T-Rex2 & 12.2 & \textbf{16.2} & 21.8 & 24.7 \\ 
    \bottomrule
    \end{tabular}%
  }
  }\\
  \subfloat[Standard fine-tuning, 4-shots]{
  \resizebox{0.48\textwidth}{!}{
    \begin{tabular}{p{2.5cm}|>{\centering\arraybackslash}p{1.5cm}>{\centering\arraybackslash}p{1.5cm}|>{\centering\arraybackslash}p{1.5cm}>{\centering\arraybackslash}p{1.5cm}}
    \toprule
    \multirow{2}[1]{*}{\textbf{Model}} & \multicolumn{2}{c|}{\textbf{Average Precsion (AP)}} & \multicolumn{2}{c}{\textbf{Average Recall (AR)}} \\
    & Chart & HRO & Chart & HRO \\
    \midrule
    \rowcolor{foundation_models!25} RegionCLIP & 6.8 & 14.7 & 15.5 & 22.9 \\
    \rowcolor{foundation_models!25} Detic & 19.6 & 14.2 & 37.0 & 22.8 \\
    \rowcolor{traditional_models!25} Faster R-CNN & 3.4 & 1.0 & 10.8 & 1.5 \\
    \rowcolor{traditional_models!25} YOLOv3 & 5.5 & 4.0 & 16.2 & 13.1 \\
    \rowcolor{traditional_models!25} RTMDet & 12.8 & 18.9 & 44.2 & 49.1 \\
    \rowcolor{traditional_models!25} Co-DETR & \textbf{27.6} & \textbf{25.5} & \textbf{53.4} & \textbf{49.7} \\
    \bottomrule
    \end{tabular}%
  }
  }
  \subfloat[Standard fine-tuning, \dataset{}]{
  \resizebox{0.48\textwidth}{!}{
    \begin{tabular}{p{2.5cm}|>{\centering\arraybackslash}p{1.5cm}>{\centering\arraybackslash}p{1.5cm}|>{\centering\arraybackslash}p{1.5cm}>{\centering\arraybackslash}p{1.5cm}}
    \toprule
    \multirow{2}[1]{*}{\textbf{Model}} & \multicolumn{2}{c|}{\textbf{Average Precsion (AP)}} & \multicolumn{2}{c}{\textbf{Average Recall (AR)}} \\
    & Chart & HRO & Chart & HRO \\
    \midrule
    \rowcolor{foundation_models!25} RegionCLIP & 10.1 & 23.3 & 17.5 & 28.6 \\
    \rowcolor{foundation_models!25} Detic & 39.6 & 34.3 & 57.4 & 47.7 \\
    \rowcolor{traditional_models!25} Faster R-CNN & 78.9 & 49.0 & 80.8 & 52.7 \\
    \rowcolor{traditional_models!25} YOLOv3 & 14.7 & 25.5 & 43.2 & 35.7 \\
    \rowcolor{traditional_models!25} RTMDet & 83.7 & 53.6 & 86.4 & 59.9 \\
    \rowcolor{traditional_models!25} Co-DETR & \textbf{88.2} & \textbf{64.0} & \textbf{89.8} & \textbf{69.5} \\
    \bottomrule
    \end{tabular}%
  }
  }
  \label{tab:detector_eval}%
  \vspace{-4mm}
\end{table}

\begin{figure}[!b]
\centering
\begin{overpic}[width=0.97\linewidth]{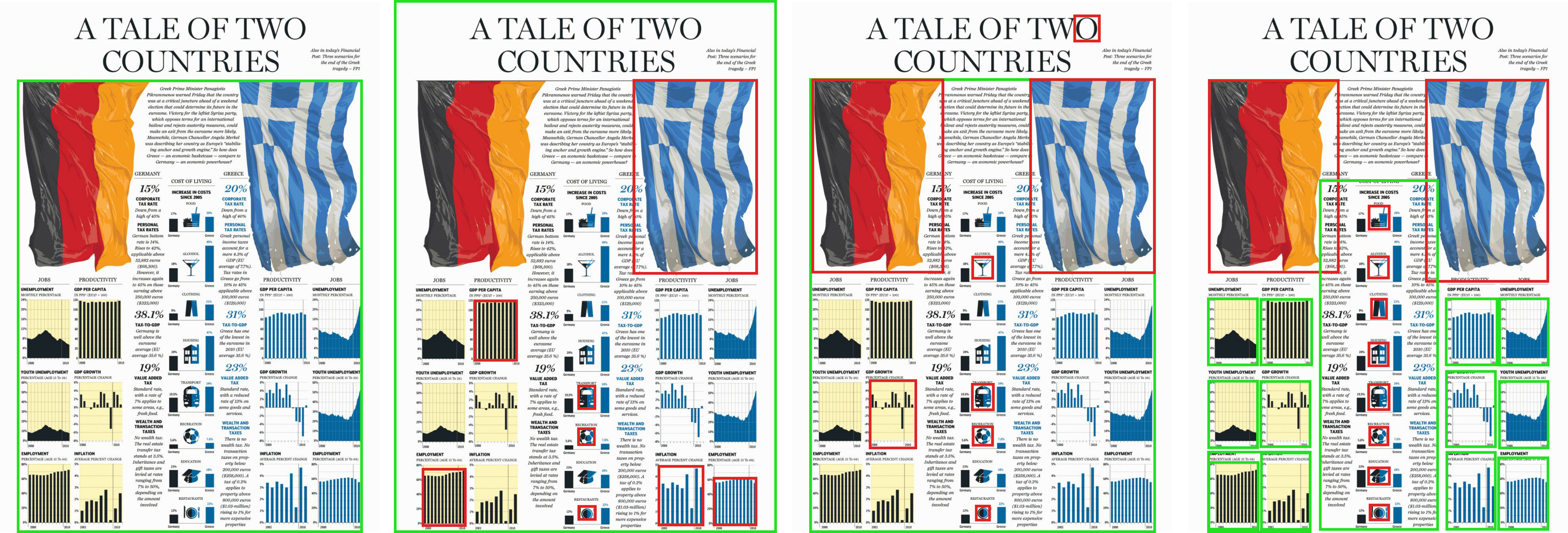}
\end{overpic}
\put(-344, -12){(a)}
\put(-247, -12){(b)}
\put(-149, -12){(c)}
\put(-52, -12){(d)}
\caption{Detection results of evaluated object detection models: (a) zero-shot prompting with DINO-X; (b) 4-shot prompting with T-Rex2; (c) 4-shot fine-tuning with Co-DETR; (d) fine-tuning on \dataset{} with Co-DETR. Bounding boxes in colors are the predictions for \fcolorbox{green}{white}{\raisebox{-0.2ex}{charts}} and 
\fcolorbox{red}{white}{\raisebox{-0.2ex}{HROs}}.}
\label{fig:det_qual}
\end{figure}

\paragraph{Comparing adaptation methods and object detection models at the element level}
We evaluate all applicable adaptation methods for each model, except for standard fine-tuning, which is restricted to models that fit within the memory constraints of an NVIDIA Tesla V100 GPU.
For few-shot prompting and fine-tuning methods, we use $k=4$, $10$, and $30$ randomly selected infographics.
We average the results over $3$ runs, excluding T-Rex2 and DINO-X due to their reliance on charged APIs.
Table~\ref{tab:detector_eval} shows the results for all models with $k=4$. 
The full results, including the variance across runs, are available in Appendix~\ref{app:full_eval}.
We present our key findings as follows:

\textit{Zero-shot and few-shot prompting exhibit limited performance}.
Zero-shot prompting exhibits limited performance.
As shown in Fig.~\ref{fig:det_qual}(a), even state-of-the-art foundation models like DINO-X fail to interpret these concepts through textual prompts, often missing key components.
Contrary to prior findings~\citep{madan2024revisiting}, providing annotated example infographics does not lead to notable improvements (Fig.~\ref{fig:det_qual}(b)).
We attribute this to the models' pretraining on natural scenes~\citep{mathew2022infographicvqa}, which provides limited exposure to graphic representations such as infographics.
Consequently, the models lack the prior knowledge needed to effectively learn from the provided examples.

\textit{Standard fine-tuning improves performance}.
Compared with zero-/few-shot prompting, fine-tuning with example infographics or the \dataset{} training set yields higher performance.
Few-shot experiments show that the performance improves significantly as the number of example infographics increases.
Moreover, fine-tuning on \dataset{} consistently outperforms few-shot fine-tuning, particularly for the traditional models.
This is verified by Co-DETR's more accurate detection results after fine-tuning on \dataset{} (Fig.~\ref{fig:det_qual}(d)) compared to using $4$ example infographics (Fig.~\ref{fig:det_qual}(c)). 

\paragraph{Evaluating traditional models at the mark level}
Due to the effectiveness of fine-tuning traditional models at the element level, we further evaluate them at the mark level.
Table~\ref{fig:fine_eval} shows the performance of each model averaged over $3$ runs.
Despite the increased difficulty of mark-level detection, which leads to a drop in mAP (\eg, from $76.1\%$ to $69.6\%$ for Co-DETR), the models still perform effectively, supported by the large-scale training set of \dataset{}.
We also evaluate model generalizability via cross-dataset evaluation following \citet{deng2022visimages}.
Specifically, we fine-tune each model on either our infographics or VG-DCU~\citep{Dou2024Hierarchically} (plain charts with mark-level annotations) and test on the other.
Models show a smaller drop in mAP when transferred from our infographics to VG-DCU than vice versa.
For example, the mAP of Co-DETR drops $17.1\%$ when transferring from our infographics to VG-DCU, compared with a much larger $53.7\%$ drop in the opposite direction.
This indicates that models trained on our infographics generalize more effectively due to the inclusion of infographic charts.
The full experimental setup and results are provided in Appendix~\ref{app:full_eval}.

\begin{table}[t]
\centering
\caption{Evaluation results of the traditional models at the element level. The best one is \textbf{bold}.}
\resizebox{0.4\linewidth}{!}{
\begin{tabular}{l|c|c}
\toprule
\textbf{Model} & \textbf{mAP} & \textbf{mAR} \\
\midrule
Faster R-CNN & 47.0 $\pm$ 0.1 & 49.9 $\pm$ 0.1 \\
YOLOv3       & 11.3 $\pm$ 0.9 & 21.5 $\pm$ 0.5\\
RTMDet       & 46.9 $\pm$ 0.2 & 59.5 $\pm$ 0.5 \\
Co-DETR      & $\mathbf{69.6\pm 0.4} $ & $\mathbf{77.0 \pm 0.3}$ \\
\bottomrule
\end{tabular}}
\vspace{-2mm}
\label{fig:fine_eval}
\end{table}

\subsection{Applying the Developed Model to Graphic Layout Detection}

To demonstrate the broader applicability of \dataset{}, we evaluate its effectiveness on graphic layout detection tasks by applying the InternImage-based model.

\subsubsection{Experimental Setup}
We evaluate the InternImage-based model on two graphic layout detection datasets, Rico~\citep{deka2017rico} and DocGenome~\citep{xia2024docgenome}.
Rico contains over 66K user interfaces collected from Android applications.
Following the common practice~\citep{manandhar2020learning, manandhar2021magic}, we aim to detect 25 UI component classes and split the dataset into 53K layouts for training and 13K for testing.
DocGenome is a large-scale scientific document dataset of 6.8M pages sourced from the arXiv repository, annotated with bounding boxes for 13 categories of components.
We randomly sample 113K pages for training and 13K for testing.
For both datasets, we fine-tune four model variants, each pre-trained on a different combination of ImageNet-22K~\citep{deng2009imagenet}, Objects365~\citep{shao2019objects365}, COCO~\citep{lin2014microsoft}, and \dataset{}, as shown in Table~\ref{tab:transfer}.
Please refer to Appendix~\ref{app:graphic_setup} for more details on the fine-tuning hyperparameters and computational costs.

\begin{table}[H]
  \vspace{-1mm}
  \centering
  \caption{Performance of the detection models with different pre-training data. The best one is \textbf{bold}.}
  \resizebox{0.65\textwidth}{!}{
    \begin{tabular}{l|c|c}
    \toprule
    \textbf{Pre-Training Data} & \textbf{Rico} & \textbf{DocGenome} \\
    \midrule
    -  & 42.1 & 69.0 \\
    \dataset{} & 50.6 & 74.4 \\ 
    \midrule
    ImageNet-22K, Objects365, COCO & 51.8 & 78.7 \\
    ImageNet-22K, Objects365, COCO, \dataset{} & \textbf{53.6} & \textbf{80.0}  \\
    \bottomrule
    \end{tabular}%
  }
  \label{tab:transfer}%
  \vspace{-1mm}
\end{table}

\subsubsection{Results and Analysis}

As shown in Table~\ref{tab:transfer}, pre-training on \dataset{} improves model performance when fine-tuned on Rico and DocGenome, both on its own and when combined with existing large-scale pre-training data.
With the growing interest in integrating multiple datasets for training foundation models~\citep{yang2023foundation}, \dataset{} serves as a useful addition to existing resources for graphic layout detection.

\section{Conclusion}
\label{sec:conclusion}

In this paper, we introduce \dataset{}, a dataset designed to support infographic element detection.
It features a diverse collection of real and synthetic infographics, along with bounding box annotations for texts, charts, HROs, and finer-grained sub-elements.
Three applications demonstrate that \dataset{} is not only valuable for developing visual reasoning methods but also broadly applicable to tasks such as object detection model evaluation and graphic layout detection.
Although \dataset{} has proven effective, there remain promising directions for future work.
For example, analyzing the annotated infographics to uncover design principles could advance automated infographic design.

\section*{Ethics Statement}

To ensure the integrity of this work, we carefully consider several ethical aspects when collecting real infographics from online platforms.
First, we manually collect the real infographics and release only their source URLs on Hugging Face, without hosting or redistributing third-party content.
We have reviewed the data usage policies of the platforms and confirmed that they either explicitly permit (\eg, Statista, Visual Capitalist) or do not prohibit (\eg, Daily Infographics, Infographics Archive) the use of their content for research purposes, including the sharing of source URLs.
Details of the data usage policies and licenses of each platform are provided in Appendix~\ref{appx:platforms}.
Second, to exclude harmful and sensitive content from our dataset, we:
1) collect only from reputable public platforms, which generally filter such content; and 
2) use GPT-4o mini to flag potentially harmful or sensitive infographics, which are then manually verified and removed.
Finally, we release our dataset and model strictly for academic research purposes.

\bibliography{reference}
\bibliographystyle{iclr2026_conference}

\newpage

\appendix

\section{Statistical Comparison of \dataset{} with existing chart datasets}

Table~\ref{tab:dataset_stats} provides a statistical comparison of \dataset{} with existing datasets.
Unlike most existing datasets that focus on plain charts, \dataset{} is specifically designed for infographic charts, where charts, texts, and HROs are tightly integrated in visually complex layouts. 
Compared to the dataset by \citet{borkin2015beyond}, the only existing infographic dataset with mark-level annotations, \dataset{} is significantly larger in scale and better suited for training object detection models.

\begin{table}[H]
\centering
\caption{Statistics of existing chart datasets.}
\begin{tabular}{l|r|r|c}
\hline
Dataset & \# Real & \# Synthetic & Infographic? \\
\hline
\dataset{} (ours) & $\numreal{}$ & $\numsyn{}$  & \checkmark \\
\citet{borkin2015beyond}     & $393$    & 0       & \checkmark \\
FigureQA~\citep{ebrahimi2018figureqa}         & 0      & $100,000$ & - \\
PlotQA~\citep{methani2020plotqa}           & 0      & $224,377$ & - \\
Beagle~\citep{battle2018beagle}            & $41,000$ & 0       & - \\
VisImages~\citep{deng2022visimages}         & $12,267$ & 0       & - \\
VG-DCU~\citep{Dou2024Hierarchically}            & $4,51$5  & $10,682$  & - \\
\hline
\end{tabular}
\label{tab:dataset_stats}
\end{table}

\section{Online Platforms for Real Infographic Collection}
\label{appx:platforms}

We collect the real infographics from the \numplat{} online platforms listed in Table~\ref{tab:infographic_platforms}.
The infographic collection strictly adheres to the copyright and licensing regulations of the respective platforms.

\begin{table}[H]
\centering
\caption{Infographic platforms and licenses.}
\label{tab:infographic_platforms}
\begin{tabular}{|L{0.25\textwidth}|L{0.25\textwidth}|C{0.40\textwidth}|}
\hline
\textbf{Platform} & \textbf{Website Link} & \textbf{Licenses} \\ \hline
Statista & \href{https://www.statista.com/}{statista.com} & \href{https://www.statista.com/getting-started/publishing-statista-content-daily-data---creative-commons}{CC BY-NC} \\ \hline
Visual Capitalist & \href{https://www.visualcapitalist.com/}{visualcapitalist.com} &  \parbox[t]{0.40\textwidth}{\href{https://licensing.visualcapitalist.com/frequently-asked-questions/}{Customized}:\\``For individuals and small organizations (5 people or less, or < $\$$1 million in revenue), we allow you to use our visuals in a variety of use cases for free. These include personal and commercial use cases, such as: embedding our graphics in a newsletter, report, video, presentation, or on your website.''} \\ \hline
World Statistics & \href{https://world-statistics.org/}{world-statistics.org} & \href{https://ourworldindata.org/faqs}{CC BY} \\ \hline
Our World in Data & \href{https://ourworldindata.org/}{ourworldindata.org} & \href{https://github.com/owid}{CC BY} \\ \hline
OECD & \href{https://www.oecd.org/}{oecd.org} & \href{https://www.oecd.org/en/about/oecd-open-by-default-policy.html}{CC BY} \\ \hline
Openverse & \href{https://openverse.org/zh-cn}{openverse.org} & \href{https://docs.openverse.org/terms_of_service.html}{CC}\\ \hline
The Conversation & \href{https://theconversation.com/}{theconversation.com} & \href{https://theconversation.com/global/terms-and-conditions}{CC BY-ND}\\ \hline
Kaiser Family Foundation & \href{https://www.kff.org/}{kff.org} & \href{https://www.kff.org/about-us/permissions-citations-reprints}{CC BY-NC-ND}  \\ \hline
Daily Infographics & \href{https://dailyinfographics.com/}{dailyinfographics.com} & \parbox[t]{0.40\textwidth}{\href{https://dailyinfographic.com/terms}{Customized}.\\ No prohibition on research use or sharing of source URLs.\\}  \\ \hline
Infographics Archive & \href{https://www.infographicsarchive.com/}{infographicsarchive.com} & \parbox[t]{0.40\textwidth}{\href{https://www.infographicsarchive.com/about/}{Customized}.\\ No prohibition on research use or sharing of source URLs.\\}  \\ \hline
\end{tabular}
\end{table}

\section{Example Synthetic Infographics}
\label{app:example}

\begin{figure}[H]
\centering
\begin{overpic}[width=0.6\linewidth]{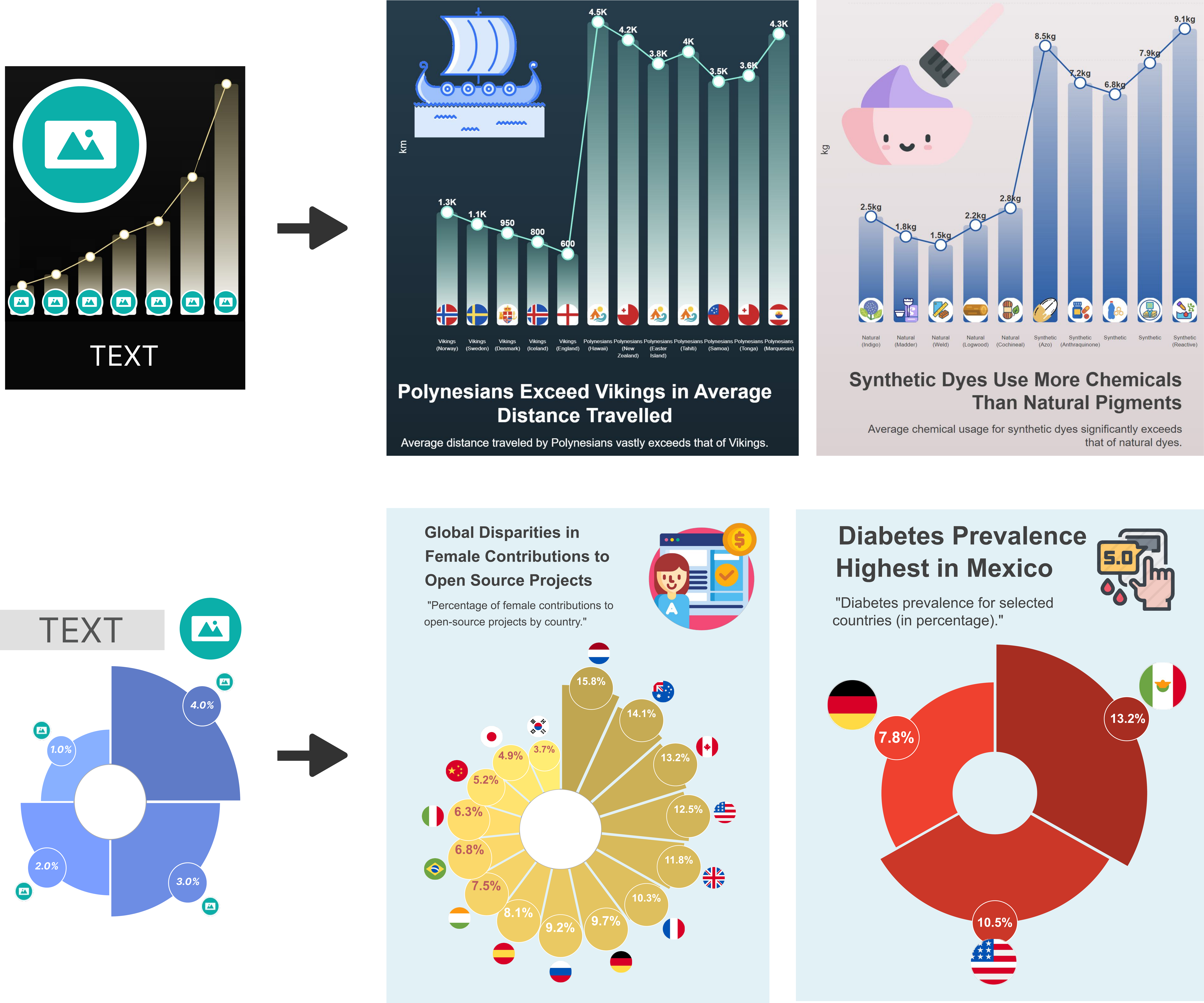}
\end{overpic}
\put(-220, -12){(a)}
\put(-81, -12){(b)}
\caption{Template-based generation of synthetic infographics: (a) design templates; (b) synthetic infographics generated from the design templates.}
\label{fig:temp_gen}
\vspace{-4mm}
\end{figure}

We employ a template-based method to create synthetic infographics.
Fig.~\ref{fig:temp_gen} shows examples of design templates and infographics generated from them.

\section{Dataset Statistics}
\label{app:stats}

\begin{figure}[H]
\centering
\includegraphics[width=1\linewidth]{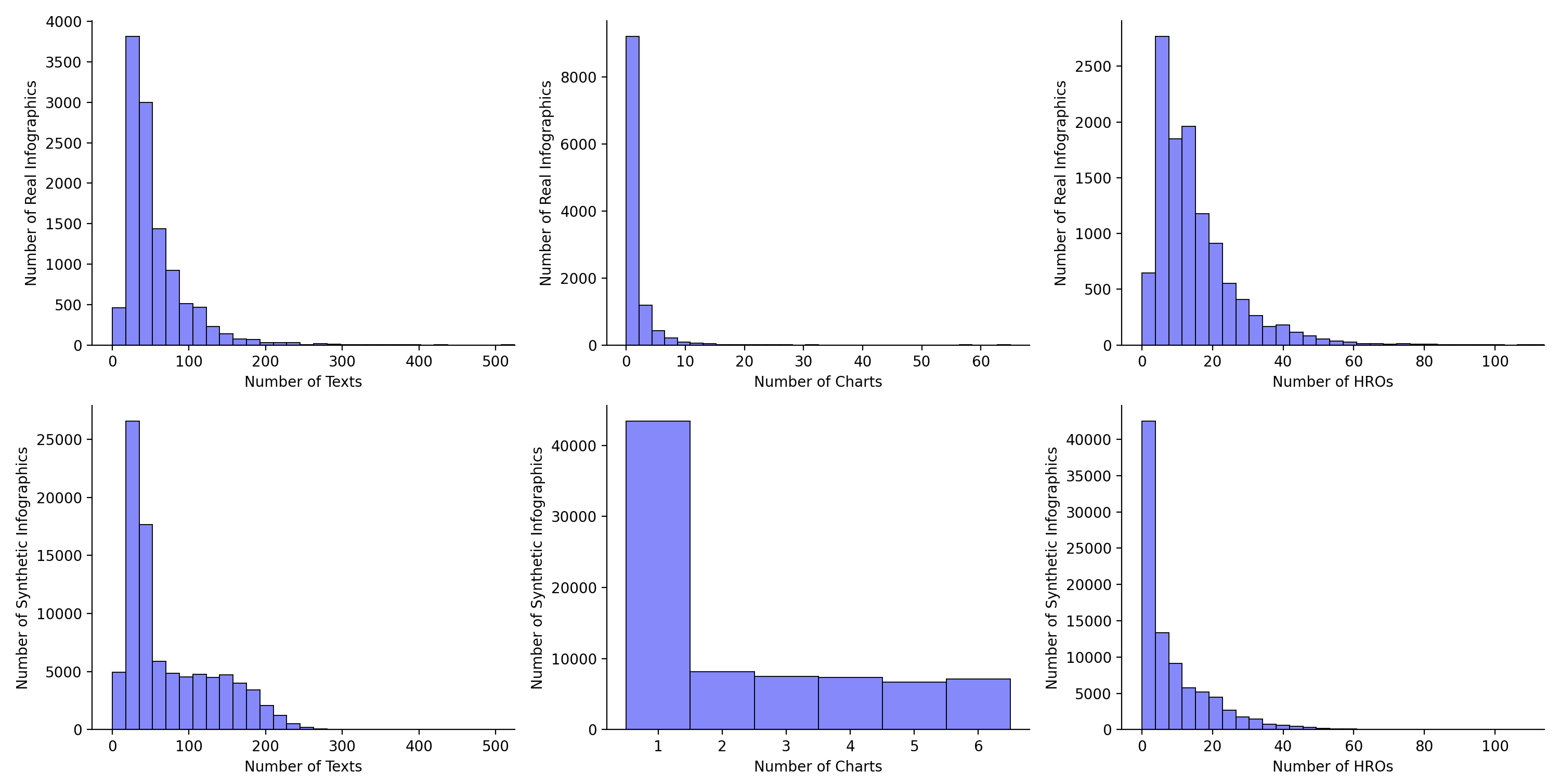}
\caption{The distribution of the number of texts, charts, and HROs in each infographic.}
\label{fig:stats_1}
\vspace{-4mm}
\end{figure}

Fig.~\ref{fig:stats_1} shows the distribution of the number of annotated texts, charts, and HROs per real and synthetic infographic.
On average, each real infographic contains $53.76$ text elements (each corresponding to a line of text), $1.94$ charts, and $14.74$ HROs, while each synthetic infographic contains $74.21$ text elements, $3.20$ charts, $10.16$ HROs, and $61.69$ sub-elements.
The slight difference in annotation density between real and synthetic infographics enhances the diversity of the dataset, improving its utility for training models to handle diverse infographics.

We classify charts and HROs into subcategories: charts are categorized into 75 distinct types, while HROs are labeled as either data-related or theme-related objects.
The $75$ chart types are:
\begin{enumerate*}[label={\arabic*)}]
    \item Vertical bar chart,
    \item Vertical stacked bar chart,
    \item Vertical grouped bar chart,
    \item Horizontal bar chart,
    \item Horizontal stacked bar chart,
    \item Horizontal grouped bar chart,
    \item Radial bar chart,
    \item Radial stacked bar chart,
    \item Radial grouped bar chart,
    \item Circular bar chart,
    \item Circular stacked bar chart,
    \item Circular grouped bar chart,
    \item Pictorial percentage bar chart,
    \item Histogram,
    \item Lollipop chart,
    \item Dot chart,
    \item Diverging bar chart,
    \item Vertical bar chart with circle,
    \item Horizontal bar chart with circle,
    \item Vertical dot bar chart,
    \item Horizontal dot bar chart,
    \item Dumbbell plot,
    \item Span chart,
    \item Bump chart,
    \item Line graph,
    \item Spline graph,
    \item Stepped line graph,
    \item Slope chart,
    \item Small multiples of line graphs,
    \item Small multiples of spline graphs,
    \item Small multiples of stepped line graphs,
    \item Area chart,
    \item Spline area chart,
    \item Layered area chart,
    \item Layered spline area chart,
    \item Range area chart,
    \item Stacked area chart,
    \item Radial area chart,
    \item Radial spline area chart,
    \item Radial layered area chart,
    \item Radial layered spline area chart,
    \item Radial range area chart,
    \item Radial stacked area chart,
    \item Diverging area chart,
    \item Diverging spline area chart,
    \item Small multiples of area charts,
    \item Small multiples of spline area charts,
    \item Pie chart,
    \item Donut chart,
    \item Semicircle pie chart,
    \item Semicircle donut chart,
    \item Rose chart,
    \item Small multiples of pie charts,
    \item Small multiples of donut charts,
    \item Small multiples of semicircle pie charts,
    \item Small multiples of semicircle donut charts,
    \item Small multiples of rose charts,
    \item Radar line chart,
    \item Radar spline chart,
    \item Small multiples of radar line charts,
    \item Small multiples of radar spline charts,
    \item Proportional area chart,
    \item Scatterplot,
    \item Grouped scatterplot,
    \item Bubble chart,
    \item Heatmap,
    \item Waffle chart,
    \item Small multiples of waffle charts,
    \item Alluvial diagram,
    \item Gauge chart,
    \item Small multiples of gauge charts,
    \item Funnel chart,
    \item Pyramid chart,
    \item Treemap,
    \item Voronoi treemap
\end{enumerate*}.

For the real infographics, we have attempted to classify the charts and HROs using GPT-4o.
However, it achieves limited accuracy, with $61.49\%$ on $1,179$ charts and $74.69\%$ on $1,498$ HROs.
As current models face challenges in reliably classifying charts and HROs in infographics, we leave their fine-grained annotation for future work.

\section{Creating Mark-Level Annotations}
\label{app:mark_cats}

To create mark-level annotations, we extend our synthetic infographic creation and annotation methods.
Using these methods, we generate annotations of $26$ element categories.
The 26 categories are: "vertical\_gridline", "dumbbell\_mark", "scatter\_mark", "legend", "bar\_mark", "proportional-area\_mark", "axis", "other\_gridline", "line\_mark", "area-under-line\_mark", "gauge\_mark", "bump\_mark", "horizontal\_gridline", "stacked-bar\_mark", "radar\_mark", "donut\_mark", "sankey\_mark", "pie\_mark", "span\_mark", "bubble\_mark", "histogram\_mark", "treemap\_mark", "waffle\_mark", "pyramid\_mark", "funnel\_mark", and "range\_mark".

\section{Dataset Analysis}
\label{app:dataset_analysis}

We have conducted analyses to verify that our dataset exhibits high diversity, no harmful bias, and high fidelity of synthetic infographics.

\subsection{Diversity Analysis}

To evaluate the diversity of the dataset, we extract a set of attributes from each infographic and examine the variety of values for each attribute. 
Specifically, we consider four key attributes: chart type, infographic topic, visual style, and communication tone. Attribute values are extracted using Gemini-2.5-flash, with slightly different strategies for each attribute: 1) Chart type is selected from the 67 identified chart categories; 2) Infographic topic was categorized based on the \citet{iptc}, a widely adopted taxonomy for news content; 3) For visual style and communication tone, the model first generate a descriptive term for each infographic, and then grouped semantically similar terms into attribute values. 
We report the number of unique values identified for each attribute in Table~\ref{tab:diversity}. 
These numbers indicate a broad range of chart types, infographic topics, visual styles, and communication tones. 
In particular, the infographic topics span all 17 top-level categories and 96 out of 120 second-level topics in the IPTC Media Topics taxonomy. 
These results verify the diversity of our dataset.

\begin{table}[H]
\centering
\caption{Diversity of infographic attribute values in \dataset{}.}
\resizebox{0.95\linewidth}{!}{%
\begin{tabular}{l|c|l}
\hline
Attribute & \# Attribute Values & Example Attribute Values \\
\hline
Chart Type & 67 & horizontal bar chart, line graph, treemap \\
Infographic Topic & 96 & public health, weather forecast, market and exchange \\
Visual Style & 15 & minimalistic, cartoonish, vintage \\
Communication Tone & 21 & neutral, critical, persuasive \\
\hline
\end{tabular}
}
\label{tab:diversity}
\end{table}

\subsection{Bias Analysis}

To mitigate potential bias, we have monitored attribute distributions during the infographic collection and generation and applied resampling techniques when necessary. 
For example, we observed an overrepresentation of bar charts in early synthetic samples and reduced their frequency through targeted resampling. 
We also computed pointwise mutual information (PMI) across attribute pairs to identify unexpected co-occurrences. Manual inspection showed that high-PMI pairs (\eg, environment topic + concerned tone) aligned with common communication patterns and did not suggest harmful or misleading bias. 
In summary, we found no harmful or systemic bias in the dataset and ensured that attribute variations remain reflective of real-world usage.

\subsection{Quality Evaluation of Synthetic Infographics and Comparison with Real Infographics}

\paragraph{Quality evaluation}
The quality of synthetic infographics is evaluated at both the sample and dataset levels. 
At the sample level, we randomly select $500$ synthetic infographics and manually verify their quality. 
All of them clearly convey the underlying data, contain HROs that align well with their intended semantics, and have accurate annotations. 
A minor issue is observed in $18$ samples, where slight unintended overlap occurs between elements due to minor misalignments during rendering. 
However, this does not affect object detection, as the boundaries of the overlapping elements remain clear. 
At the dataset level, we evaluate how well the synthetic infographics cover the real ones in feature space. Specifically, we use CLIP~\citep{radford2021learning} to extract the feature embeddings of all infographics and project them into a two-dimensional space using UMAP~\citep{mcinnes2018umap}. 
The space is divided into a uniform grid, and a real infographic is considered ``covered" if at least one synthetic infographic falls in the same grid cell. 
The results show that 92.64\% of the real infographics are covered, indicating the high representativeness of the synthetic infographics.

\begin{figure}[H]
\centering
\begin{overpic}[width=0.5\linewidth]{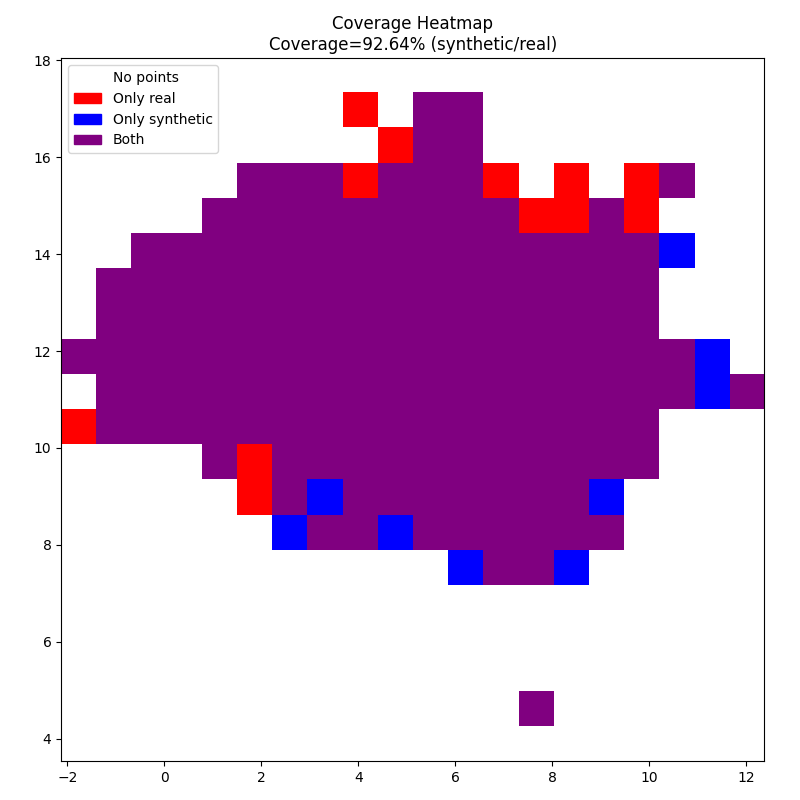}
\end{overpic}
\caption{The distribution of synthetic and real infographics.}
\label{fig:umap}
\vspace{-4mm}
\end{figure}

\paragraph{Comparison with real infographics}
We qualitatively examine the difference in the distribution of synthetic and real infographics in the UMAP visualization (Fig.~\ref{fig:umap}). 
While the overall distribution of synthetic and real infographics largely overlaps, some distinctive characteristics are observed. 
Synthetic infographics include chart variations specially crafted by design experts, such as those with hand-drawn-style fills, which are relatively rare in real infographics. 
Real infographics, on the other hand, uniquely feature composite charts, where multiple single charts are combined by sharing axes or overlaying each other. 

\section{Detailed Experimental Setup}

\subsection{Thinking-with-Boxes via Grounded Chain-of-Thought}
\label{app:GCoT_setup}

\paragraph{Prompts for the grounded chain-of-thought method and the baselines}
In the grounded chain-of-thought method, we prepend the grounded infographic elements to the question-category-specific prompt used in ChartQAPro~\citep{masry2025chartqapro}.
Below is an example input to the vision-language model.

\begin{tcolorbox}[colframe=cyan,colback=cyan!5,title=Example Prompt for Grounded Chain-of-Thought, breakable]
{
    \fontsize{8pt}{10pt} \selectfont
    You will be provided with two versions of the same infographic chart, each with certain elements highlighted. \\
    You will also be provided with the information lists of elements highlighted in the images. Each entry in the lists of elements follows the format (ID, Content), where: \\
    \hspace*{2em}ID means the id of the element. \\
    \hspace*{2em}Content means the content of the element.
    
    \vspace{10pt}
    
    \begin{minipage}{1\textwidth}
        \centering
        \includegraphics[height=200pt]{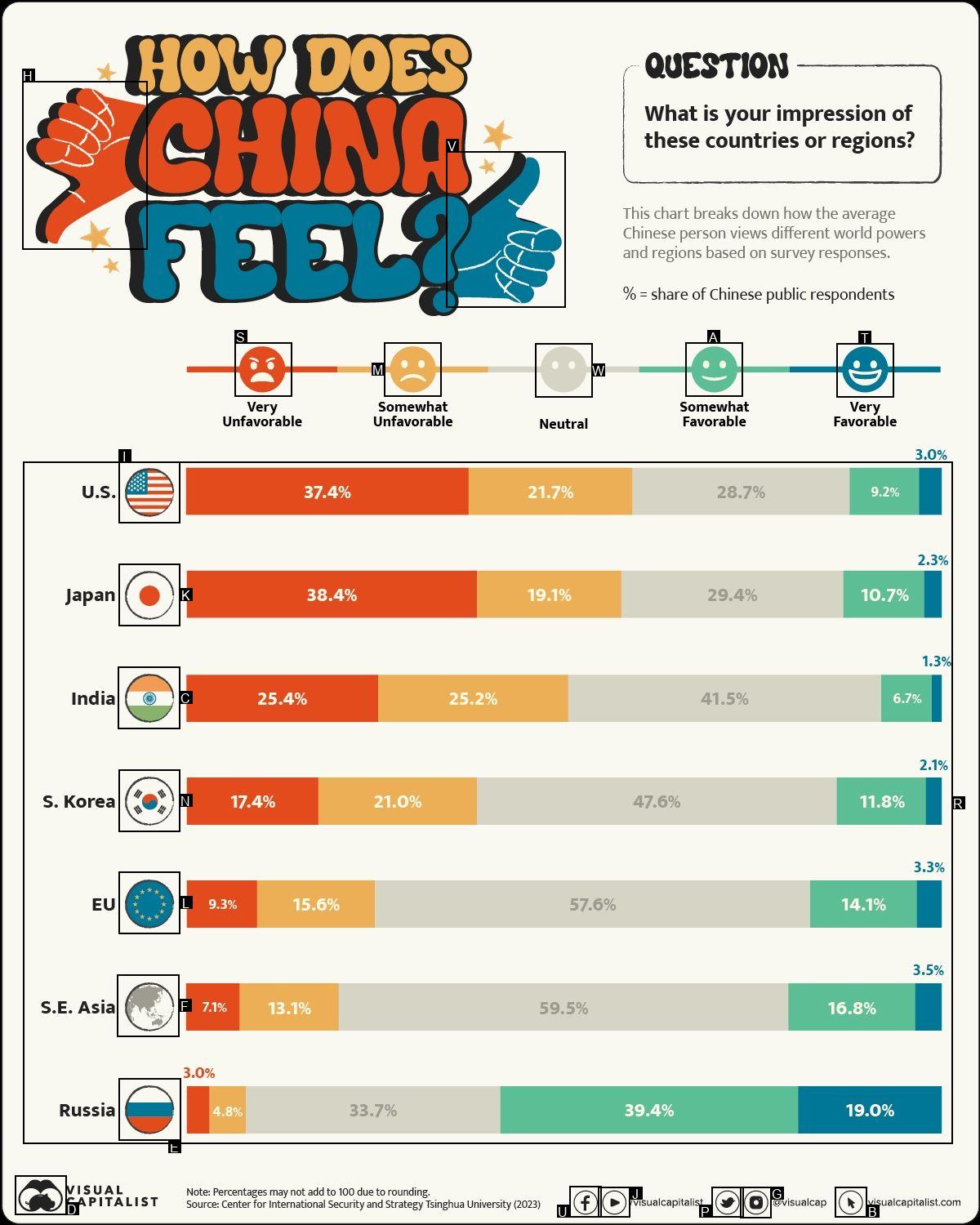}\\
    \end{minipage}


    This above image highlights non-text elements enclosed in boxes, each labeled with a unique ID. \\
    Here is the list of elements: \\
    *************************************** \\
    (ID=A, Content="human recognizable object") \\
    ...... \\
    (ID=R, Content="chart") \\
    ...... \\
    ***************************************
    
    \vspace{10pt}
    
    \begin{minipage}{1\textwidth}
        \centering
        \includegraphics[height=200pt]{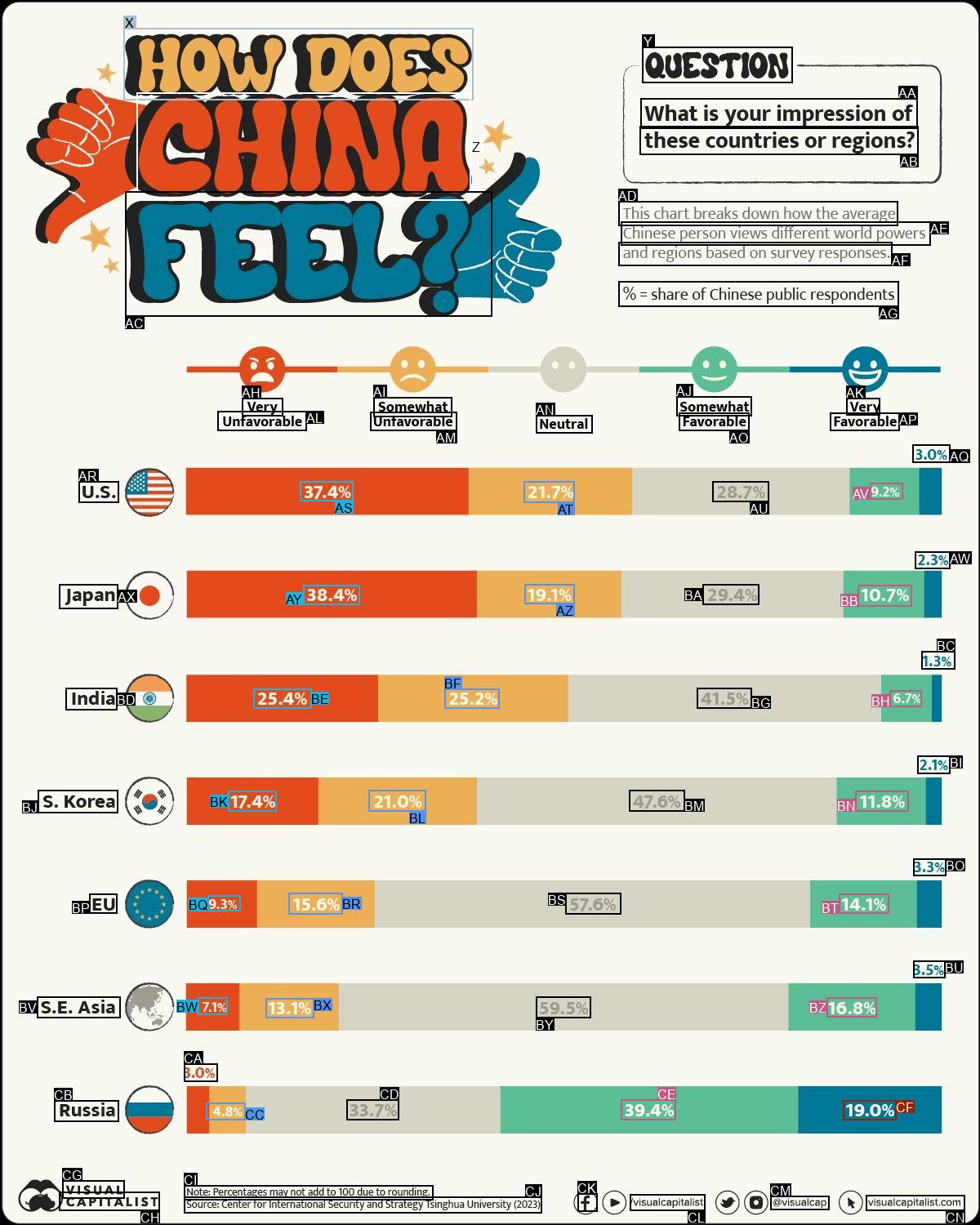}\\
    \end{minipage}
    
    
    This above image highlights text elements enclosed in boxes, each labeled with a unique ID. \\
    Here is the list of elements: \\
    *************************************** \\
    (ID=X, Content="text:~~HOW DOES") \\
    (ID=Y, Content="text:~~QUESTION") \\
    ...... \\
    *************************************** \\
    \\
    These labeled elements are intended to support you in your upcoming task. Please refer to and make use of them as needed during your thinking and analysis, and be sure to mention their IDs when doing so. \\
    For example: \\
    1. Based on the content in box ID 1, (your finding about the box), or; \\
    2. Based on the relationships of box ID 1, ID 2, ..., ID N, (your finding based on the boxes). \\
    \\
    Below is the image of original infographic chart, followed by your task:
    
    
    \begin{minipage}{1\textwidth}
        \centering
        \includegraphics[height=200pt]{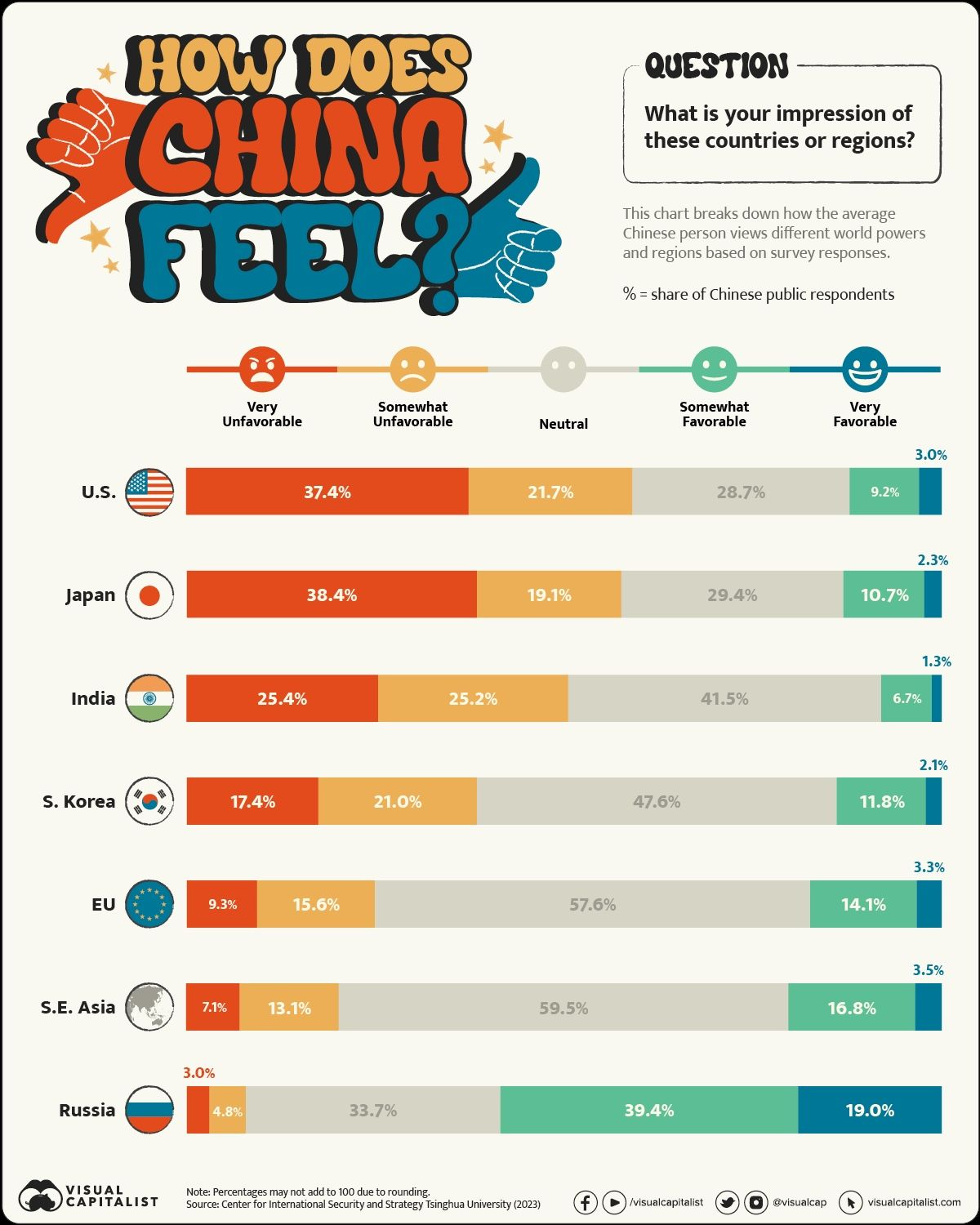}\\
    \end{minipage}
    
    
    You are given a factoid question that you need to answer based on the provided image.
    \\
    You need to think step-by-step, but your final answer should be a single word, number, or phrase. If the question is unanswerable based on the information in the provided image, your answer should be unanswerable. Do not generate units. But if numerical units such as million, m, billion, B, or K are required, use the exact notation shown in the chart.
    \\
    If there are multiple final answers, put them in brackets using this format ['Answer1', 'Answer2'].
    \\
    Remember to think step-by-step and mention the IDs of the elements you used, and reply in the following JSON format:
    \\
    \{
    \\
    \hspace*{2em}"Steps": "The step-by-step thinking process with IDs mentioned.",
    \\
    \hspace*{2em}"A": "Your answer."
    \\
    \}
    \\
    \textcolor{blue!60}{\textbf{Question:} }{\setlength{\spaceskip}{0.3em} \ttfamily What proportion of Chinese public respondents have a neutral impression of Japan?}

}
\end{tcolorbox}

For the baselines, we use the same prompt as ChartQAPro.
Below are examples of the input for the three baselines: direct prompting, chain-of-thought, and program-of-thought.

\begin{tcolorbox}[colframe=cyan,colback=cyan!5,title=Example Prompt for Direct Prompting, breakable]
{
    \fontsize{8pt}{10pt} \selectfont
    \begin{minipage}{1\textwidth}
        \centering
        \includegraphics[height=200pt]{figs/example_ori.jpg}\\
    \end{minipage}
    
    
    You are given a factoid question that you need to answer based on the provided image.
    \\
    Your answer should be a single word, number, or phrase. If the question is unanswerable based on the information in the provided image, your answer should be unanswerable. Do not generate units. But if numerical units such as million, m, billion, B, or K are required, use the exact notation shown in the chart.
    \\
    If there are multiple final answers, put them in brackets using this format ['Answer1', 'Answer2'].
    \\
    Remember to generate the final answer only without any additional text!
    \\
    \textcolor{blue!60}{\textbf{Question:} }{\setlength{\spaceskip}{0.3em} \ttfamily What proportion of Chinese public respondents have a neutral impression of Japan?}
}
\end{tcolorbox}

\begin{tcolorbox}[colframe=cyan,colback=cyan!5,title=Example Prompt for Chain-of-Thought, breakable]
{
    \fontsize{8pt}{10pt} \selectfont
    \begin{minipage}{1\textwidth}
        \centering
        \includegraphics[height=200pt]{figs/example_ori.jpg}\\
    \end{minipage}
    
    
    You are given a factoid question that you need to answer based on the provided image.
    \\
    You need to think step-by-step, but your final answer should be a single word, number, or phrase. If the question is unanswerable based on the information in the provided image, your answer should be unanswerable. Do not generate units. But if numerical units such as million, m, billion, B, or K are required, use the exact notation shown in the chart.
    \\
    If there are multiple final answers, put them in brackets using this format ['Answer1', 'Answer2'].
    \\
    Remember to think step-by-step and format the final answer in a separate sentence like "The answer is X"
    \\
    \textcolor{blue!60}{\textbf{Question:} }{\setlength{\spaceskip}{0.3em} \ttfamily What proportion of Chinese public respondents have a neutral impression of Japan?}
}
\end{tcolorbox}

\begin{tcolorbox}[colframe=cyan,colback=cyan!5,title=Example Prompt for Program-of-Thought, breakable]
{
    \fontsize{8pt}{10pt} \selectfont
    \begin{minipage}{1\textwidth}
        \centering
        \includegraphics[height=200pt]{figs/example_ori.jpg}\\
    \end{minipage}
    
    
    You are given a factoid question that you need to answer based on the provided image.
    \\
    You need to write an executable python code that calculates and prints the final answer, but your final answer should be a single word, number, or phrase. If the question is unanswerable based on the information in the provided image, your answer should be unanswerable. Do not generate units. But if numerical units such as million, m, billion, B, or K are required, use the exact notation shown in the chart.
    \\
    If there are multiple final answers, put them in brackets using this format ['Answer1', 'Answer2'].
    \\
    Remember to return a python code only without any additional text.
    \\
    \textcolor{blue!60}{\textbf{Question:} }{\setlength{\spaceskip}{0.3em} \ttfamily What proportion of Chinese public respondents have a neutral impression of Japan?}
}
\end{tcolorbox}

\begin{figure}[H]
\centering
\centering
\begin{overpic}[width=0.99\linewidth]{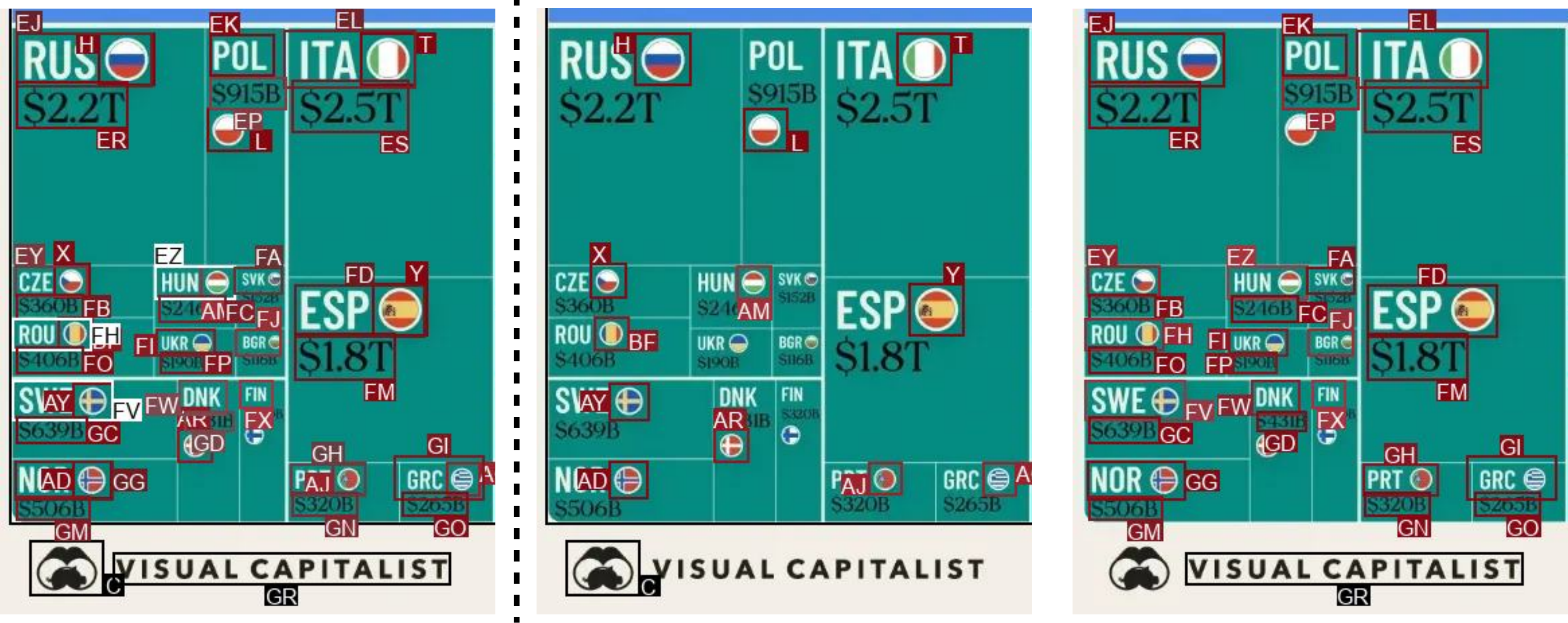}
\end{overpic}
\put(-332, -12){(a)}
\put(-134, -12){(b)}
\caption{Comparison of the visual prompts rendered with different layer configurations: (a) visual prompts rendered in one merged layer: (b) visual prompts rendered in two separate layers.}
\label{fig:layers}
\vspace{-4mm}
\end{figure}

\paragraph{Comparison of the visual prompts rendered in one layer versus two layers}
In our grounded chain-of-thought method, we propose to separate the visual prompts into two layers: one for charts and HROs, and the other for texts.
As shown in Fig.~\ref{fig:layers}, this separation improves visual clarity by reducing overlap between bounding boxes.

\paragraph{Enhanced relaxed accuracy implementation}
Following ChartQAPro, we use the enhanced relaxed accuracy to evaluate the chart understanding performance.
This metric evaluates answers based on the following criteria:
\begin{enumerate}
    \item Numeric answers are allowed a 5\% error margin.
    \item For answers in `years', an exact match is required.
    \item Textual answers are evaluated using the ANLS score~\citep{biten2019scene}, which is based on the edit distance between texts.
    \item Multiple-choice and fact-checking tasks are evaluated using an exact-match criterion.
\end{enumerate}
To more accurately evaluate model performance, we make three refinements to the official implementation of the enhanced relaxed accuracy:
\begin{enumerate}
    \item We remove punctuation marks (\emph{i.e.}, commas and periods) from answers, ensuring that `25,000' and `25000' are treated as equivalent.
    \item We remove unit symbols when evaluating numeric answers, so that values like `100' and `\$100' are treated as equivalent.
    \item We standardize ratios and percentages by converting them into decimal form, so that expressions like `3:2', `150\%', and `1.5' are all treated as equivalent.
\end{enumerate}

\subsection{Comparing Object Detection Models}
\label{app:benchmark_setup}

We evaluate the performance of existing object detection models in detecting charts and HROs.
As the models are not tailored to detecting charts and HROs, we adapt them using three adaptation methods:
\begin{enumerate*}[label={\arabic*)}]
    \item \textbf{Zero-shot prompting}, which uses text prompts to define target classes,
    \item \textbf{Few-shot prompting}, which uses $k$ randomly selected infographics to describe target classes, optionally augmented with text prompts, and
    \item \textbf{Standard fine-tuning}, which updates model weights using annotated infographics, either with $k$ random example infographics or the \dataset{} training set.
\end{enumerate*}

For zero-shot prompting, we evaluate six models: RegionCLIP~\citep{zhong2022regionclip}, Detic~\citep{zhou2022detecting}, Grounding DINO~\citep{liu2024grounding}, GLIP~\citep{li2022grounded}, MQ-GLIP~\citep{xu2023multi}, and DINO-X~\citep{ren2024dinoxunifiedvisionmodel}, all of which take the class names "chart" and "human recognizable object" as input.

For few-shot prompting, we evaluate two models: T-Rex2~\citep{jiang2024t} and MQ-GLIP~\citep{xu2023multi}.
For T-Rex2, we provide $k$ randomly selected infographics with bounding box annotations.
For MQ-GLIP, we provide the class names along with the selected infographics.

For traditional fine-tuning, we evaluate six models: RegionCLIP, Detic, Faster R-CNN~\citep{ren2015faster}, YOLOv3~\citep{redmon2018yolov3}, RTMDet~\citep{lyu2022rtmdet}, and Co-DETR~\citep{zong2023detrs}.
For fine-tuning on the entire \dataset{} training set, we train for $E$ epochs with a batch size of $\mathcal{B}$ and a learning rate of $lr$.
Table~\ref{tab:continual_param} shows the fine-tuning hyperparameters, which adhere to the official settings, as well as the computational costs, in terms of GPU hours using NVIDIA GeForce RTX 4090 D.
For few-shot fine-tuning, we adjust the number of training epochs inversely with the number of random infographics, ensuring consistent computational costs.
All other fine-tuning hyperparameters remain unchanged.

\begin{table}[H]
  \renewcommand{\arraystretch}{1.1} 
  \centering
  \caption{Training hyperparameters and computational costs for traditional fine-tuning on the entire \dataset{} training set.}
  \resizebox{0.95\textwidth}{!}{
    \begin{tabular}{c|c|c|c|c|c|c}
    \toprule
    Hyperparameters & RegionCLIP & Detic & Faster R-CNN & YOLOv3 & RTMDet & Co-DETR \\
    \midrule
    Optimizer & SGD & AdamW& SGD & SGD & AdamW & AdamW \\
    $E$ & 1 & 8 & 10 & 10 & 5 & 3 \\
    $\mathcal{B}$ & 1 & 8 & 64 & 64 & 64 & 64 \\
    $lr$ & $5e-4$ & $3.75e-6$ & $2e-3$ & $1e-3$ & $4e-3$ & $1e-5$ \\
    \makecell{Computational costs\\(GPU hours)} & 20 & 40 & 20 & 30 & 40 & 70 \\
    \bottomrule
    \end{tabular}%
  }
  \label{tab:continual_param}%
\end{table}%

\subsection{Applying the Developed Model to Graphic Layout Detection}
\label{app:graphic_setup}

We evaluate the InternImage-based model on two graphic layout detection datasets, Rico~\citep{deka2017rico} and DocGenome~\citep{xia2024docgenome}.
Rico contains over 66K user interfaces collected from Android applications.
Following the common practice~\citep{manandhar2020learning, manandhar2021magic}, we aim to detect 25 UI component classes and split the dataset into 53K layouts for training and 13K for testing.
DocGenome is a large-scale scientific document dataset of 6.8M pages sourced from the arXiv repository, annotated with bounding boxes for 13 categories of components.
We randomly sample 113K pages for training and 13K for testing.
Following the official setting~\citep{wang2023internimage}, we fine-tune the frozen InternImage backbones along with the DINO detector~\citep{zhang2023dino} for $12$ epochs.
The batch size is set to $16$, and we use an AdamW optimizer~\citep{loshchilov2018decoupled} with an initial learning rate of 0.0001 and a weight decay of 0.05. 
We use a step-based learning rate scheduler, which decreases the learning rate by a factor of 0.1 at epochs 8 and 11.
The training takes 196 GPU hours on Rico and 296 GPU hours on DocGenome using NVIDIA Tesla V100.

\section{Detailed Analysis of errors by o3 on ChartQAPro}
\label{app:GCoT_ana}

Despite its strong visual reasoning capability, o3 achieves slightly lower accuracy compared to o1 and o4-mini on the ChartQAPro benchmark~\citep{masry2025chartqapro}.
To investigate this, we randomly sample $200$ question-answer pairs and analyze the failure patterns when using grounded CoT.
We identify two primary sources of failures:
\begin{enumerate*}[label={\arabic*)}]
    \item \textbf{perception error}, where models fail to correctly interpret the content and relationships of the infographic elements, and
    \item \textbf{instruction following error}, where models do not  adhere to the prompt when formatting the answer
\end{enumerate*}.
As shown in Table~\ref{tab:error_analysis}, perception errors are the main cause of chart understanding failures, occurring with similar frequency across all models.
However, o3 shows a higher frequency of instruction-following errors, contributing to its slightly lower overall performance compared to o1 and o4-mini.
In particular, even when instructed to output the numerical answer as a single word, o3 often includes extra words like `$\approx$' and `about'.
To address this, we have attempted to increase the reasoning effort from ‘medium’ to ‘high’.
However, as shown in Table~\ref{tab:ablate_effort}, this change does not yield an obvious improvement in the chart understanding performance, and the instruction following error still occurs with a similar frequency.
This suggests that the `medium' setting already provides sufficient reasoning budget for ChartQAPro, and alternative strategies are needed to enhance o3's instruction-following ability.

\begin{table}[H]
  \vspace{-4mm}
  \centering
  \caption{Error analysis of chart understanding failures on ChartQAPro for o1, o3, and o4-mini.}
  \resizebox{0.5\textwidth}{!}{
    \renewcommand{\arraystretch}{1.0} 
    \begin{tabular}{l|cc}
    \toprule
    \textbf{Model} & \textbf{Perception Error} &  \textbf{\makecell{Instruction\\Following Error}} \\
    \midrule
    o1 & 48 & 12 \\ 
    o3 & 47 & 22 \\
    o4-mini & 46 & 8 \\
    \bottomrule
    \end{tabular}
  }
  \label{tab:error_analysis}%
\end{table}

\begin{table}[H]
  \vspace{-4mm}
  \centering
  \caption{Performance of o3 using different levels of reasoning effort.}
  \resizebox{0.65\textwidth}{!}{
    \renewcommand{\arraystretch}{1.0} 
    \begin{tabular}{l|cccc}
    \toprule
     \textbf{Reasoning Effort} & \textbf{Direct} & \textbf{CoT} & \textbf{PoT} & \textbf{Grounded CoT} \\
    \midrule
    Medium & 60.6 & 60.0 & 59.5 & 61.6 \\
    High & 60.4 & 61.0 & 60.8 & 61.8 \\
    \bottomrule
    \end{tabular}
  }
  \label{tab:ablate_effort}%
\end{table}

\section{Detailed Evaluation Results}
\label{app:full_eval}

\paragraph{Comparing adaptation methods and object detection models}
We evaluate all applicable adaptation methods for each model, except for standard fine-tuning, which is restricted to models that fit within the memory constraints of an Nvidia Tesla V100 GPU.
For few-shot prompting and fine-tuning methods, we use $k=4$, $10$, and $30$ randomly selected infographics.
We average the results over $3$ runs, excluding T-Rex2 and DINO-X, due to their reliance on charged APIs.
Tables~\ref{tab:detector_eval_AP} and \ref{tab:detector_eval_AR} show the AP and AR along with their standard deviation for all models.

\begin{table}[H]
  \centering
  \caption{AP of object detection models for the chart and HRO categories. The best one is \textbf{bold}.}
  \resizebox{0.99\textwidth}{!}{
    \renewcommand{\arraystretch}{1.0} 
    \begin{tabular}{ll|c|ccc|cccc}
    \toprule
    \multicolumn{2}{c|}{\multirow{2}[1]{*}{\textbf{Model}}}  & \multirow{2}[1]{*}{\textbf{\makecell{Zero-shot\\prompting}}} & \multicolumn{3}{c|}{\textbf{\makecell{Few-shot prompting}}} & \multicolumn{4}{c}{\textbf{\makecell{Standard fine-tuning}}} \\
    & & & 4-shots & 10-shots & 30-shots & 4-shots & 10-shots & 30-shots & \dataset{} \\
    \midrule
    \multicolumn{10}{c}{\large{\textbf{Chart Category}}} \\
    \midrule
    \multirow{7}{*}{\rotatebox{90}{\makecell{Foundation\\Models}}} 
    & RegionCLIP & 0.83 & - & - & - & 6.81 $\pm$ 3.70 & 9.38 $\pm$ 1.62 & 10.12 $\pm$ 0.96 & 10.14 $\pm$ 0.80 \\
    & Detic & 1.77 & - & - & - & 19.62 $\pm$ 5.61 & 23.02 $\pm$ 1.38  & 28.00 $\pm$ 1.42 & 39.57 $\pm$ 0.10 \\
    & Grounding Dino & 12.64 & - & - & - & - & - & - & - \\
    & GLIP & 13.52 & - & - & - & - & - & - & - \\
    & MQ-GLIP & 13.52 & \textbf{16.24 $\pm$ 0.63} & \textbf{16.80 $\pm$ 0.19} & \textbf{16.91 $\pm$ 0.24} & - & - & - & - \\
    & T-Rex2 & - & 12.22 & - & - & - & - & - & - \\
    & DINO-X & \textbf{13.99} & - & - & - & - & - & - & - \\
    \midrule
    \multirow{4}{*}{\rotatebox{90}{\makecell{Traditional\\Models}}} 
    & Faster R-CNN & - & - & - & - & 3.43 $\pm$ 2.46 & 5.63 $\pm$ 2.34 & 9.89 $\pm$ 2.89 & 78.92 $\pm$ 0.33 \\
    & YOLOv3 & - & - & - & - & 5.49 $\pm$ 1.64 & 5.12 $\pm$ 2.81 & 7.83 $\pm$ 2.04 & 14.70 $\pm$ 5.92 \\
    & RTMDet & - & - & - & - & 12.82 $\pm$ 1.74 & 25.90 $\pm$ 2.44 & 28.84 $\pm$ 2.09 & 83.65 $\pm$ 3.46 \\
    & Co-DETR & - & - & - & - & \textbf{27.63 $\pm$ 11.41} & \textbf{31.65 $\pm$ 4.20} & \textbf{43.36 $\pm$ 2.99} & \textbf{88.22 $\pm$ 0.64} \\
    \midrule
    \multicolumn{10}{c}{\large{\textbf{HRO Category}}} \\
    \midrule
    \multirow{7}{*}{\rotatebox{90}{\makecell{Foundation\\Models}}} 
    & RegionCLIP & 3.57 & - & - & - & 14.70 $\pm$ 0.55 & 18.28 $\pm$ 0.23 & 18.80 $\pm$ 1.38 & 23.26 $\pm$ 0.31 \\
    & Detic & 4.38 & - & - & - & 14.24 $\pm$ 5.15 & 22.47 $\pm$ 3.29 & 30.41 $\pm$ 1.19 & 34.31 $\pm$ 0.59 \\
    & Grounding Dino & 12.24 & - & - & - & - & - & - & - \\
    & GLIP & 11.21 & - & - & - & - & - & - & - \\
    & MQ-GLIP & 11.21 & 15.46 $\pm$ 1.43 & \textbf{16.18 $\pm$ 0.36} & \textbf{16.94 $\pm$ 0.29} & - & - & - & - \\
    & T-Rex2 & - & \textbf{16.15} & - & - & - & - & - & - \\
    & DINO-X & \textbf{14.94} & - & - & - & - & - & - & - \\
    \midrule
    \multirow{4}{*}{\rotatebox{90}{\makecell{Traditional\\Models}}} 
    & Faster R-CNN & - & - & - & - & 0.98 $\pm$ 0.13 & 4.07 $\pm$ 1.05 & 11.59 $\pm$ 1.01 &  49.02 $\pm$ 0.17 \\
    & YOLOv3 & - & - & - & - & 3.96 $\pm$ 0.77 & 5.95 $\pm$ 1.52 & 9.14 $\pm$ 1.28 & 25.52 $\pm$ 2.48 \\
    & RTMDet & - & - & - & - & 18.91 $\pm$ 2.41 & 21.91 $\pm$ 1.34 & 26.38 $\pm$ 3.29 & 53.64 $\pm$ 0.22 \\
    & Co-DETR & - & - & - & - & \textbf{25.46 $\pm$ 2.43} & \textbf{31.09 $\pm$ 1.10} & \textbf{36.94 $\pm$ 3.62} & \textbf{64.02 $\pm$ 4.73} \\
    \bottomrule
    \end{tabular}
  }
  \label{tab:detector_eval_AP}%
  \vspace{-4mm}
\end{table}

\begin{table}[H]
  \centering
  \caption{AR of object detection models for the chart and HRO categories. The best one is \textbf{bold}.}
  \resizebox{0.99\textwidth}{!}{
    \renewcommand{\arraystretch}{1.0} 
    \begin{tabular}{ll|c|ccc|cccc}
    \toprule
    \multicolumn{2}{c|}{\multirow{2}[1]{*}{\textbf{Model}}}  & \multirow{2}[1]{*}{\textbf{\makecell{Zero-shot\\prompting}}} & \multicolumn{3}{c|}{\textbf{\makecell{Few-shot prompting}}} & \multicolumn{4}{c}{\textbf{\makecell{Standard fine-tuning}}} \\
    & & & 4-shots & 10-shots & 30-shots & 4-shots & 10-shots & 30-shots & \dataset{} \\
    \midrule
    \multicolumn{10}{c}{\large{\textbf{Chart Category}}} \\
    \midrule
    \multirow{7}{*}{\rotatebox{90}{\makecell{Foundation\\Models}}} 
    & RegionCLIP & 13.93 & - & - & - & 15.50 $\pm$ 3.22 & 19.37 $\pm$ 1.47 & 19.54 $\pm$ 0.81 & 17.48 $\pm$ 0.65 \\
    & Detic & 23.72 & - & - & - & 36.99 $\pm$ 6.02 & 40.40 $\pm$ 0.97 & 44.06 $\pm$ 1.39 & 57.38 $\pm$ 0.41 \\
    & Grounding Dino & \textbf{63.22} & - & - & - & - & - & - & - \\
    & GLIP & 44.89 & - & - & - & - & - & - & - \\
    & MQ-GLIP & 44.88 & \textbf{43.47 $\pm$ 0.46} & \textbf{43.81 $\pm$ 0.51} & \textbf{43.97 $\pm$ 0.22} & - & - & - & - \\
    & T-Rex2 & - & 21.84 & - & - & - & - & - & - \\
    & DINO-X & 29.36 & - & - & - & - & - & - & - \\
    \midrule
    \multirow{4}{*}{\rotatebox{90}{\makecell{Traditional\\Models}}} 
    & Faster R-CNN & - & - & - & - & 10.79 $\pm$ 4.69 & 15.86 $\pm$ 2.74 & 20.86 $\pm$ 4.79 & 80.84 $\pm$ 0.31 \\
    & YOLOv3 & - & - & - & - & 16.21 $\pm$ 2.54 & 15.98 $\pm$ 5.08 & 23.33 $\pm$ 1.55 & 43.16 $\pm$ 5.45 \\
    & RTMDet & - & - & - & - & 44.16 $\pm$ 0.67 & 53.48 $\pm$ 4.74 & 56.34 $\pm$ 1.92 & 86.41 $\pm$ 0.25 \\
    & Co-DETR & - & - & - & - & \textbf{53.41 $\pm$ 12.34} & \textbf{61.10 $\pm$ 6.36} & \textbf{68.36 $\pm$ 2.80} & \textbf{89.82 $\pm$ 0.53} \\
    \midrule
    \multicolumn{10}{c}{\large{\textbf{HRO Category}}} \\
    \midrule
    \multirow{7}{*}{\rotatebox{90}{\makecell{Foundation\\Models}}} 
    & RegionCLIP & 24.92 & - & - & - & 22.85 $\pm$ 1.55 & 26.21 $\pm$ 0.65 & 27.08 $\pm$ 0.43 & 28.56 $\pm$ 0.27 \\
    & Detic & 11.31 & - & - & - & 22.80 $\pm$ 6.66 & 34.52 $\pm$ 5.80 & 46.16 $\pm$ 1.44 & 47.74 $\pm$ 0.50 \\
    & Grounding Dino & \textbf{45.97} & - & - & - & - & - & - & - \\
    & GLIP & 33.21 & - & - & - & - & - & - & - \\
    & MQ-GLIP & 33.20 & \textbf{40.72 $\pm$ 2.06} & \textbf{41.69 $\pm$ 1.45} & \textbf{42.38 $\pm$ 0.54} & - & - & - & - \\
    & T-Rex2 & - & 24.72 & - & - & - & - & - & - \\
    & DINO-X & 29.14 & - & - & - & - & - & - & - \\
    \midrule
    \multirow{4}{*}{\rotatebox{90}{\makecell{Traditional\\Models}}} 
    & Faster R-CNN & - & - & - & - & 1.55 $\pm$ 1.01 & 8.77 $\pm$ 3.56 & 25.92 $\pm$ 2.03 & 52.69 $\pm$ 0.21 \\
    & YOLOv3 & - & - & - & - & 13.06 $\pm$ 1.76 & 18.75 $\pm$ 1.03 & 25.31 $\pm$ 0.98 & 35.70 $\pm$ 2.01 \\
    & RTMDet & - & - & - & - & 49.10 $\pm$ 0.69 & 49.72 $\pm$ 0.94 & 52.50 $\pm$ 1.15 & 59.94 $\pm$ 0.14 \\
    & Co-DETR & - & - & - & - & \textbf{49.74 $\pm$ 0.35} & \textbf{57.15 $\pm$ 0.38} & \textbf{61.48 $\pm$ 0.84} & \textbf{69.47 $\pm$ 0.92} \\
    \bottomrule
    \end{tabular}
  }
  \label{tab:detector_eval_AR}%
  \vspace{-4mm}
\end{table}

\paragraph{Cross-dataset evaluation between our dataset and VG-DCU}
We evaluate the generalizability of the traditional models via cross-dataset evaluation following \citet{deng2022visimages}.
Specifically, we train each model on either our infographics or VG-DCU~\citep{Dou2024Hierarchically}, which comprises plain charts with element-level annotations, and evaluate each model on the other dataset.
To support the transfer between datasets, we identify common categories with identical annotation guidelines, resulting in four categories: "bar\_mark", "line\_mark", "pie\_mark", and "axis".
We train and evaluate the models on these categories.
Table~\ref{tab:cross_dataset_results} shows the evaluation results.
Models show a smaller drop in mAP when transferred from our infographics to VG-DCU compared to the opposite direction.
This shows that models trained on our data exhibit stronger generalizability due to the inclusion of infographic charts.

\begin{table}[H]
\centering
\caption{Cross-dataset evaluation results}
\begin{tabular}{l|cc|c|cc|c}
\hline
Training set & \multicolumn{3}{c|}{VG-DCU} & \multicolumn{3}{c}{Ours} \\
\cline{2-7}
Test set& VG-DCU & Ours & mAP$\downarrow$ & Ours & VG-DCU & mAP$\downarrow$ \\
\hline
Faster R-CNN & 59.5 & 19.2 & 40.3 & 53.7 & 19.1 & \textbf{34.6} \\
YOLOv3       & 26.4 & 4.1  & 22.3 & 23.2 & 12.5 & \textbf{10.7} \\
RTMDet       & 70.2 & 28.1 & 42.1 & 60.9 & 42.4 & \textbf{18.5} \\
Co-DETR      & 86.7 & 33.0 & 53.7 & 71.0 & 53.9 & \textbf{17.1} \\
\hline
\end{tabular}
\label{tab:cross_dataset_results}
\end{table}

\section{The Use of Large Language Models}

The use of LLMs in this work is limited to the following: 1) polishing the writing for clarity; 2) filtering candidate infographics during dataset construction.

\end{document}